\documentclass[conference]{IEEEtran}
\IEEEoverridecommandlockouts
\usepackage{cite}
\usepackage{amsmath,amssymb,amsfonts}
\usepackage{algorithmic}
\usepackage{graphicx}
\usepackage{textcomp}
\usepackage{xcolor}
\usepackage{multirow}
\usepackage{booktabs}
\usepackage{tabularx}
\usepackage{subfigure}
\usepackage{stfloats}

\def\BibTeX{{\rm B\kern-.05em{\sc i\kern-.025em b}\kern-.08em
    T\kern-.1667em\lower.7ex\hbox{E}\kern-.125emX}}
\begin{document}

\title{LogGPT: Exploring ChatGPT for Log-Based Anomaly Detection}
\author{Jiaxing Qi,
        Shaohan Huang,
        Zhongzhi Luan,
        Carol Fung,
        Hailong Yang,
        Depei Qian 
\IEEEcompsocitemizethanks{\IEEEcompsocthanksitem Jiaxing Qi, Shaohan Huang, Zhongzhi Luan, Hailong Yang, and Depei Qian are with the Sino-German Joint Software Institute, Beihang University, Beijing, 100191, China.\protect\\
E-mail: \{jiaxingqi, luan.zhongzhi\}@buaa.edu.cn.
\IEEEcompsocthanksitem Carol Fung is with the Concordia Institute for Information Systems Engineering, Concordia University, Quebec, Canada.}
\thanks{(Zhongzhi Luan is the corresponding author for this work.)}
}


\maketitle

\begin{abstract}
The increasing volume of log data produced by software-intensive systems makes it impractical to analyze them manually. Many deep learning-based methods have been proposed for log-based anomaly detection. These methods face several challenges such as high-dimensional and noisy log data, class imbalance, generalization, and model interpretability. Recently, ChatGPT has shown promising results in various domains. However, there is still a lack of study on the application of ChatGPT for log-based anomaly detection. In this work, we proposed LogGPT, a log-based anomaly detection framework based on ChatGPT. By leveraging the ChatGPT's language interpretation capabilities, LogGPT aims to explore the transferability of knowledge from large-scale corpora to log-based anomaly detection. We conduct experiments to evaluate the performance of LogGPT and compare it with three deep learning-based methods on BGL and Spirit datasets. LogGPT shows promising results and has good interpretability. This study provides preliminary insights into prompt-based models, such as ChatGPT, for the log-based anomaly detection task. 
\end{abstract}


\section{Introduction}
Log-based anomaly detection is an important technique to monitor system activities and identify suspicious behaviors. Logs contain the records of various operations, events, and status information, which are critical for troubleshooting and security analysis. However, manually analyzing large volumes of logs is impractical. In recent years, many automated log-based anomaly detection methods have been proposed, including rule-based methods \cite{breier2017dynamic}, machine learning (ML)-based methods \cite{breier2015anomaly}, and deep learning (DL)-based methods. Among these methods, deep learning-based methods \cite{du2017deeplog, meng2019loganomaly} have shown supreme performance. They are capable of learning complex patterns and representations from the logs, which allows them to effectively identify anomalies that may not be detected by traditional rule-based or ML-based methods \cite{he2021survey}.
\par
Deep learning-based methods leverage the power of neural networks and advanced techniques such as recurrent neural networks (RNNs) \cite{du2017deeplog}, convolutional neural networks (CNNs) \cite{lu2018detecting}, and Transformer \cite{huang2020hitanomaly} to effectively capture complex patterns and dependencies in system logs. Despite these advantages, challenges still present, such as high-dimensional and noisy log data, class imbalance, generalization across datasets, and the interpretability of models. 
\par
Recently, large language models (LLMs), such as ChatGPT \cite{cheng2023gpt}, have shown promising results in many domains such as language understanding \cite{frieder2023mathematical}, dialogue \cite{chen2022would} and machine translation \cite{jiao2023chatgpt}.  Liu et al. \cite{liu2023chatgpt} leverage ChatGPT to recommend products for users. Cheng et al. \cite{cheng2023gpt} used GPT-4 to perform end-to-end data analysis with databases from a wide range of domains. Le et al. \cite{le2023evaluation} leverage ChatGPT as a log parser to extract the log event and parameters. Nonetheless, log-based anomaly detection with ChatGPT has not been thoroughly investigated. To the best of our knowledge, this is the first study on how to use ChatGPT in log-based anomaly detection.
\par
In this work, we propose LogGPT, a log-based anomaly detection framework based on ChatGPT, which consists of three components: \textit{log preprocessing, prompt construction, and response parser}. The objective of LogGPT is to utilize ChatGPT's ability on language understanding from large-scale corpora in the domain of system log analysis. The \textit{log preprocessing} component involves filtering, parsing, and grouping to transform raw log messages into a structured format for further analysis. \textit{Prompt construction} focuses on designing specific prompts tailored to log anomaly detection, aiming to instruct ChatGPT's generation process toward accurate anomaly identification. The \textit{response parser} is responsible for extracting the output returned by ChatGPT, allowing for further analysis and evaluation of the detected anomalies. 
\par
LogGPT utilizes its language generation capabilities for log anomaly detection. We investigate the possibility to transfer the knowledge and patterns learned by ChatGPT from diverse textual sources to the specialized domain of system log analysis, enabling the effective detection of abnormal events. By conducting comprehensive experiments and analyses, we aim to gain a deeper understanding of the potential and limitations of ChatGPT for log-based anomaly detection. 
Particularly, we focus on answering the following research questions: \textit{1) What is the current capability of ChatGPT for log-based anomaly detection? and  2) How explainable are the anomalies detected by the model? } 
\par
To answer the above questions, we conduct a systematic evaluation of LogGPT and compare it to three deep learning-based methods (including DeepLog \cite{du2017deeplog}, LogAnomaly \cite{meng2019loganomaly}, LogRobust \cite{zhang2019robust}) on two datasets (BGL and Spirit \cite{le2022log}), under controlled experimental settings. We first investigate the impact of different variables (including prompt construction and window size) on LogGPT performance. Then, we compare the performance of LogGPT to three baseline methods. Finally, we explore the interpretability of LogGPT. Through extensive experiments, we obtained the following major findings about the LogGPT for log-based anomaly detection: 
\begin{itemize}
    \item The prompt construction (both task description and human knowledge injection) has a significant impact on LogGPT. A more specific task description and injecting normal log information are often beneficial.
    \item The window size affects the performance of LogGPT. Increasing the window size usually results in better performance. 
    \item Compared to three deep learning-based methods, LogGPT shows promising performance (zero-shot and few-shot) on both the BGL and Spirit datasets.
    \item LogGPT demonstrates excellent interpretability in detecting anomalies, providing users with specific information to aid in understanding the causes of anomalies and offering potential preventive suggestions.
\end{itemize}
In summary, the major contributions of this work are as follows:
\begin{itemize}
    \item This study represents the first attempt to employ ChatGPT for log-based anomaly detection and provides a quantitative evaluation of its effectiveness.
    \item We designed LogGPT, a common framework specifically designed for log-based anomaly detection. The LogGPT framework combines three components to identify and analyze anomalies within log data.
    \item We conducted extensive experiments on both BGL and Spirit datasets and demonstrated that LogGPT has promising performance and good interpretability.
\end{itemize}

\begin{figure}[!tb]
    \centering
    \includegraphics[scale=0.55]{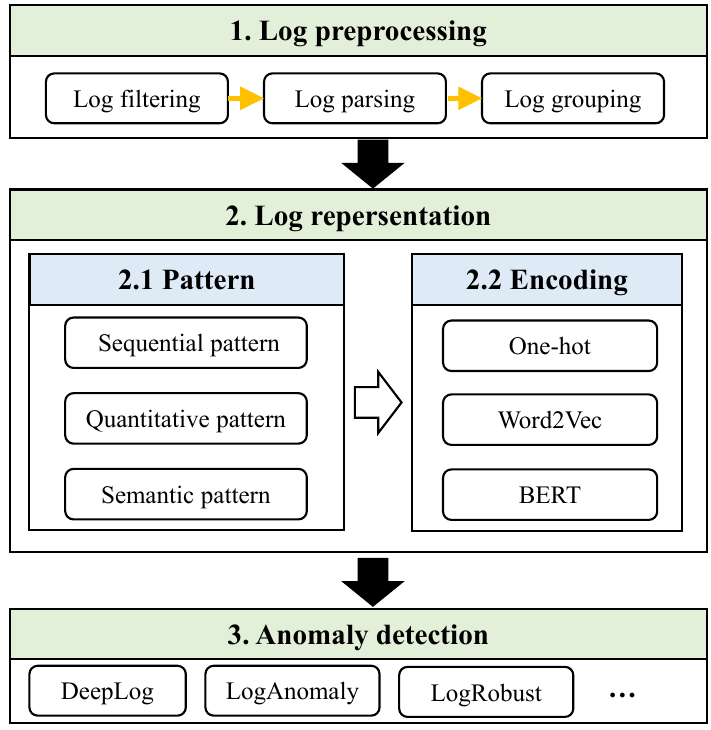}
    \caption{Log-based anomaly detection workflow.}
    \label{fig:workflow}
\end{figure}
\par

\section{Preliminary}
The common workflow of log-based anomaly detection is shown in Figure \ref{fig:workflow}, which includes three steps: \textit{1)} log preprocessing, \textit{2)} log representation, \textit{3)} anomaly detection through DL models.

\subsubsection{Log preprocessing} Log messages are semi-structured texts, which consist of a constant part (log event) and a variable part (parameters). Generally, we require to filter, parse, and group raw logs to train an anomaly detection model. First, some noise is removed by log filtering. Then, log events and parameters are automatically extracted using log parsing methods. For example, many log parsing methods have been proposed, such as Drain \cite{he2017drain}, Spell \cite{du2016spell}, and Paddy \cite{huang2020paddy}. Finally, logs will be separated into various groups, where each group contains several log records. These groups are called log sequences, which will extract various patterns as the input of anomaly detection models.  
\subsubsection{Log representation} Log sequences necessitate transformation into feature vectors for utilization as input of DL models. There are three primary types of log patterns: \textbf{Sequential pattern}, which represents the contextual information of log sequences. \textbf{Quantitative pattern}, which statistic each log event occurs distribution within log sequences. \textbf{Semantic pattern}, which represents the semantic meaning of each log event using a language model, aims to extract the associated semantic information of log sequences. In order to represent the aforementioned log patterns in the form of a feature vector, commonly employed encoding methods include one-hot encoding, word2vec, and BERT \cite{le2022log}.
\subsubsection{Anomaly detection}  The main purpose of this step is to train a deep anomaly detection model with input as feature vectors of log sequences. A variety of DL techniques have been applied to log-based anomaly detection, such as CNN, RNN, and Transformer \cite{le2022log}. These models can be grouped into three types based on training strategy: supervised, semi-supervised, and unsupervised. Supervised models consider log-based anomaly detection as a binary classification task, which utilizes both normal and abnormal logs in the training stage. Semi-supervised models capture normal patterns from the normal log sequences to detect anomalies. Unsupervised models do not require any labeled logs and typically combine with cluster and generative methods.


\section{Log-based Anomaly Detection with ChatGPT}
\begin{figure}[!tb]
    \centering
    \includegraphics[scale=0.6]{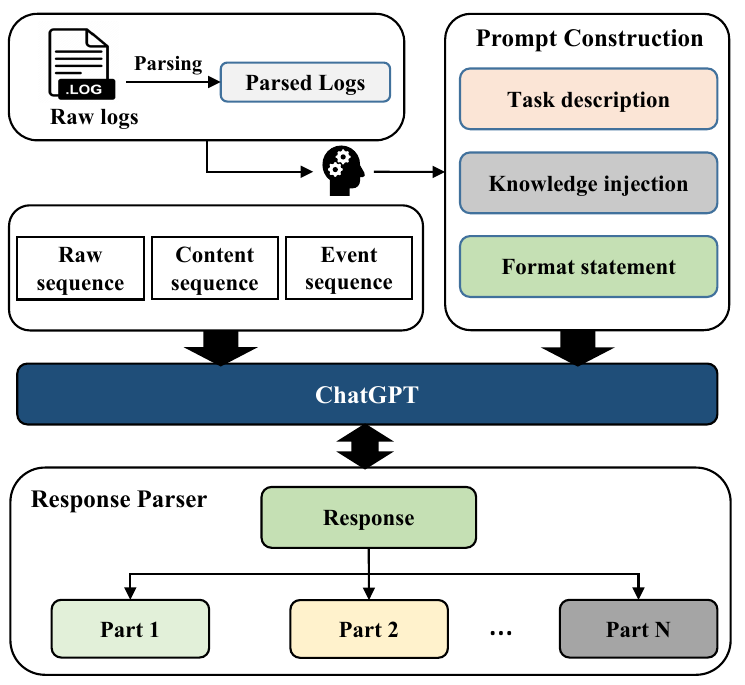}
    \caption{The framework of LogGPT to perform log-based anomaly detection.}
    \label{fig:framework}
\end{figure}

We design a framework, namely LogGPT, for log-based anomaly detection using ChatGPT. As shown in Figure \ref{fig:framework}, the framework consists of three main components: \textbf{Log preprocessing}, where the raw log messages are parsed into a structured format; \textbf{Prompt construction}, where different anomaly detection prompts are constructed for log sequences; and \textbf{Response parser}, where prompts and sequences will form a request to be sent to ChatGPT. The response information will be parsed into parts for evaluation, and the final result will be presented to the users.

\begin{figure}
    \centering
    \includegraphics[scale=0.6]{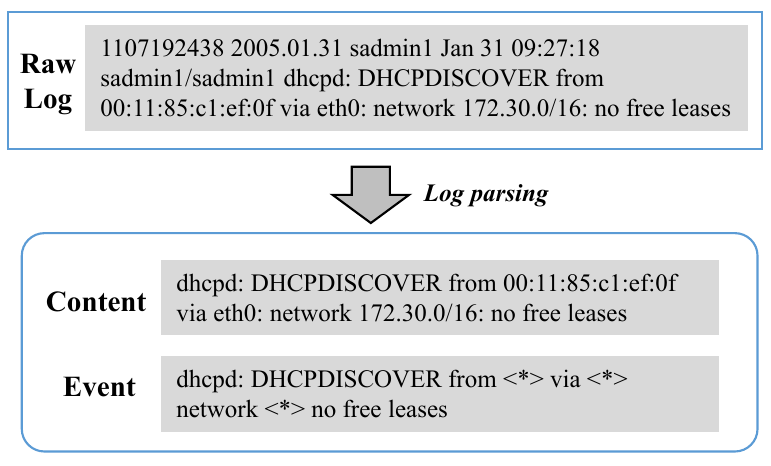}
    \caption{An example of \textit{log parsing}.}
    \label{fig:logparse}
\end{figure}

\subsection{Log Preprocessing}
This step is to extract the structured information from raw logs. We employed the state-of-the-art log parsing method (Drain) to extract structured data, including \textit{ID}, \textit{Timestamp}, \textit{Content}, and \textit{EventTemplate}. After that, the raw logs and parsed logs are grouped into different chunks using a fixed-size time window, which generates three types of sequences: \textit{raw sequence, content sequence, and event sequence}. An example of this step is shown in Figure \ref{fig:logparse}.
\par
The \textit{raw sequence} consists of the raw log messages, capturing the unaltered information directly from the log messages. The \textit{content sequence} focuses on the log text. It excludes certain content that is irrelevant to the analysis, such as \textit{ID} and \textit{Timestamp}. The event sequence is the high-level abstraction of the \textit{content sequence} where dropped the variable part in the log text. Compared to the three types of sequence, the \textit{raw sequence} provides fine-grained information, enabling detailed investigation and troubleshooting. The \textit{content sequence} emphasizes the textual content, facilitating text-based analysis and anomaly detection. The event sequence provides coarse-grained information, highlighting sequential patterns to support pattern-based anomaly detection.
 
\subsection{Prompt Construction}
\begin{figure*}[!tb]
    \centering
    \includegraphics[scale=0.66]{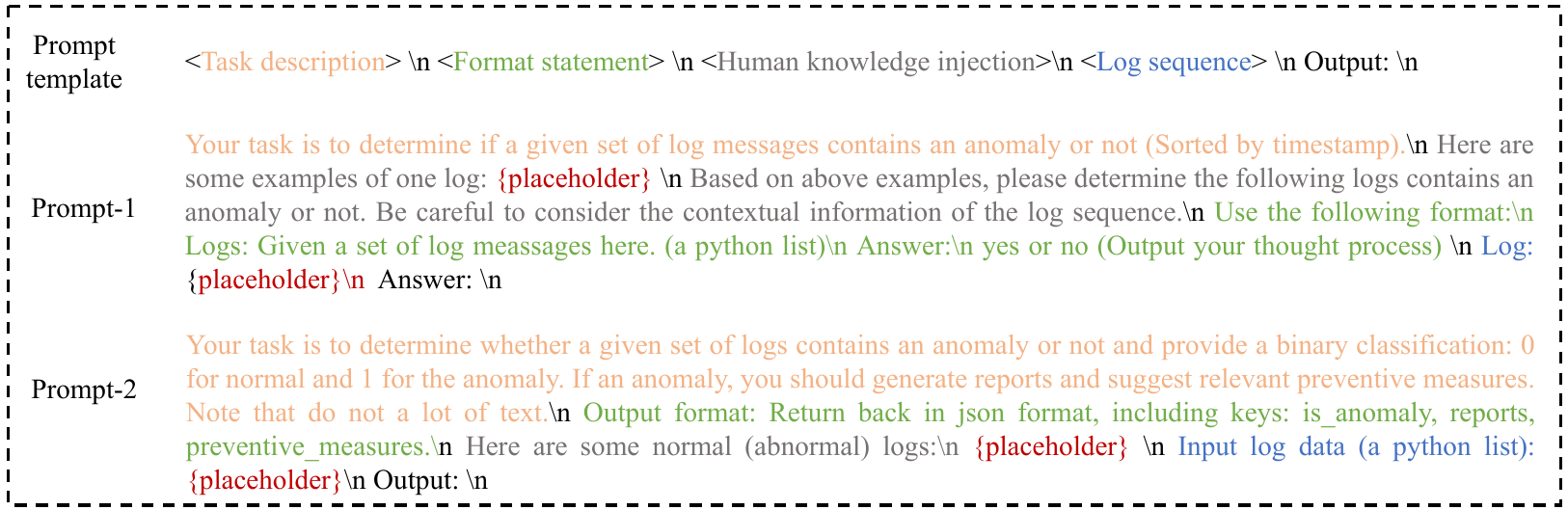}
    \caption{Example prompts of log-based anomaly detection. For the zero-shot setting, the human-knowledge injection part is dropped.}
    \label{fig:prompt_template}
\end{figure*}

In this step, we describe the prompt construction strategy for the log-based anomaly detection task. As shown in Figure \ref{fig:prompt_template}, we designed a prompt template, which includes the following parts: \textit{task description}, \textit{format statement}, \textit{human knowledge injection}, \textit{input sequence}. Then, we fill in the content of each part by domain experience and improve the prompt using ChatGPT. In addition, we follow the general tips that include more specific, starting simple and iterating on improvements, for designing task-specific prompts. Finally, we selected two prompts with the best test performance to conduct our experiments. 
\par

1) \textit{Task description.} Log-based anomaly detection is a crucial task for maintaining system operations. While DL-based methods have made some progress in this field, their interpretability remains a challenging issue \cite{he2021survey}. Therefore, the \textit{task description} should not only instruct ChatGPT to determine whether an anomaly has occurred but also prompt it to provide explanations for the occurrence of anomalies. Furthermore, it should guide ChatGPT to suggest possible preventive measures. To achieve these objectives, we introduce two types of instruction in this part, namely indirect instruction and direct instruction. For indirect instruction, we use a phrase, such as \textit{"Output your thought process"}, to prompt ChatGPT to explain the anomalous events. For direct instruction, we explicitly instruct ChatGPT to generate anomaly reports and preventive measures. This way, we aim to provide users with more fine-grained system reports and help users achieve a deeper understanding of the operation status of the system.
\par
2) \textit{Format statement.} To ensure response diversity, ChatGPT incorporates some randomness during the generation process. Additionally, when users sent requests to ChatGPT, they can specify a temperature parameter to control the diversity of the response. Higher temperature values make ChatGPT prefer to select words and phrases more randomly, resulting in more diverse and creative text generation \cite{liu2023chatgpt}. However, higher temperature values may also cause the model to choose less common or less reasonable vocabulary, leading to unexpected responses. While such responses can be more creative, they also lead to inaccurate or unreasonable outcomes, making it challenging to evaluate the performance of anomaly detection. To address this concern, we introduce two methods to handle this challenge. First, we explicitly indicate the expected response format in the prompt. For example, we indicate the response must be in json format and specify the keys that should be included, which ensures that most of the responses meet the expected format (\textit{Output format: Please note that return back in following json format, include keys: is\_anomaly, reports, preventive\_measures}). However, there may still be some responses that do not meet the expectations. We introduce the \textit{Response Parser} component (see Section \ref{sec:response_parser}) to address this issue.
\par
3) \textit{Human Knowledge Injection.} DL-based methods have demonstrated that the performance is often unsatisfactory without any domain prior knowledge (unsupervised methods) \cite{he2021survey}. Therefore, it should be beneficial to introduce some prior knowledge in the prompt. We have introduced this part (optional, gray) to the prompt template to allow users to inject specific domain prior knowledge into the prompt, which aims to improve the performance of ChatGPT on log anomaly detection tasks. Generally, we refer to this kind of prompt as a few-shot setting, otherwise, it is a zero-shot setting. For example, we can use some labeled logs to fill this part, so that ChatGPT can preview some domain knowledge.
\par
4) \textit{Input sequence.}  There are three types of input sequences mentioned above. To this end, we provide a variable part (blue) that supports different input sequences. When sending a request to ChatGPT, we form the log sequence into a Python list and concatenate other parts as a final prompt. 

\subsection{Response Parser} \label{sec:response_parser}
The responses of ChatGPT have diversity. In addition, it should be noted that the output token of ChatGPT has a maximum length limit for each request. Once the limit is exceeded, the response will be terminated and the current output will be returned.  To ensure the parsability of the response, we designed this component to parse the text of the response into several pre-defined parts. For example, three parts are defined in this study including \textit{is\_anomaly, reports, and preventive\_measures}. For each response, we format and check the response text. If the pre-defined format is satisfied,  the text is parsed directly. Otherwise, we use ChatGPT to reformat the text so that it satisfies the requirements.

\section{Evaluation}
In this section, we evaluate LogGPT by answering the following research questions:
\begin{itemize}
    \item \textbf{RQ1:} How does LogGPT perform with different prompts?
    \item \textbf{RQ2:} How does LogGPT perform on different window sizes?
    \item \textbf{RQ3:} How does human knowledge injection affect the performance of LogGPT?
    \item \textbf{RQ4:} How does LogGPT perform compared to deep learning-based methods?
    \item \textbf{RQ5:} How explainable are the anomalies detected by LogGPT?
\end{itemize}
\subsection{Datasets}
Our performance evaluation process is based on two commonly used log datasets \cite{le2022log}, and the details of each dataset are as follows:
\begin{itemize}
    \item \textbf{Blue Gene/L (BGL)}  dataset contains 4,747,963 log messages from a Blue Gene/L supercomputer system at Lawrence Livermore National Laboratory, California. It has 131,072 processors and 32,768GB of memory. The dataset includes alert and non-alert messages, with 348,460 (7.34\%) labeled as anomalous.
    \item \textbf{Spirit} dataset is collected from a Linux production cluster at Sandia National Labs, comprising 512 nodes over a 23-day period. It contains 272,298,969 log messages, with 172,816,564 (63.47\%) labeled as system anomalies. 
\end{itemize}
\par
We split the training and testing set with 8:2. Due to the imposed limitations on ChatGPT's API request frequency, we adopted a random sampling strategy to select a subset consist 2000 consecutive logs from the testing set. Moreover, we have checked manually to ensure a proportion of abnormal logs within the subset. This approach aimed to sample a representative subset of logs for evaluation.
\subsection{Baseline methods and Metrics}
\begin{itemize}
    \item \textbf{Deeplog \cite{du2017deeplog}} is a semi-supervised method that uses sequential vectors as input patterns to learn normal system executions by predicting the next log event based on preceding events.
    \item \textbf{LogAnomaly \cite{meng2019loganomaly}} is a semi-supervised method that uses sequential vector and quantitative vector as input, as well as applies an LSTM-based model to detect sequential and quantitative anomalies in log sequence.
    \item \textbf{LogRobust \cite{zhang2019robust}} is a supervised log-based anomaly detection method that extracts semantic information of log events and represents them as semantic vectors. It then uses an attention-based Bi-LSTM model to detect anomalies, which can capture contextual information in log sequences and have better robust for log unstable.
\end{itemize}
\par
\textbf{Evaluation Metrics.} Like previous works \cite{le2022log,zhang2022deeptralog}, we use common evaluation metrics for anomaly detection tasks, including F1 score, Precision, and Recall. In addition, Specificity is also used in order to evaluate more comprehensively.
\subsection{Implemention details} 
To extract the contents and events from the log data, we use the log parser \textit{Drain} with the default parameter settings \cite{he2021survey}. Note that the special character (\textit{$\langle*\rangle$}) in log events is dropped when grouped into event sequences. For the ChatGPT, we apply \textit{gpt-3.5-turbo}\footnote{https://platform.openai.com/docs/models/overview} to conduct zero-shot and few-shot experiments with two prompts (Figure \ref{fig:prompt_template}). We set the temperature is 0 and only the top-1 choice is returned. Moreover, the maximum number of output tokens is limited to 100 for a faster response. In the few-shot setting, we put the 5 historical logs and labels to enable ChatGPT to learn some prior knowledge. For the response parser, we designed a prompt to reformat the response in \textit{json} format if the pre-defined format is not satisfied: \textit{"Please format the following text in json format, which include the keys: ... "}. The implementation of three baseline methods is referred to the public code on GitHub\footnote{https://github.com/LogIntelligence/LogADEmpirical}.

\subsection{Results and Analysis}

\subsubsection{\textbf{RQ1: Performance with different prompts}}  
\
\par
\begin{figure*}[!tb]
    \centering
    \subfigure[BGL]{\includegraphics[scale=0.085]{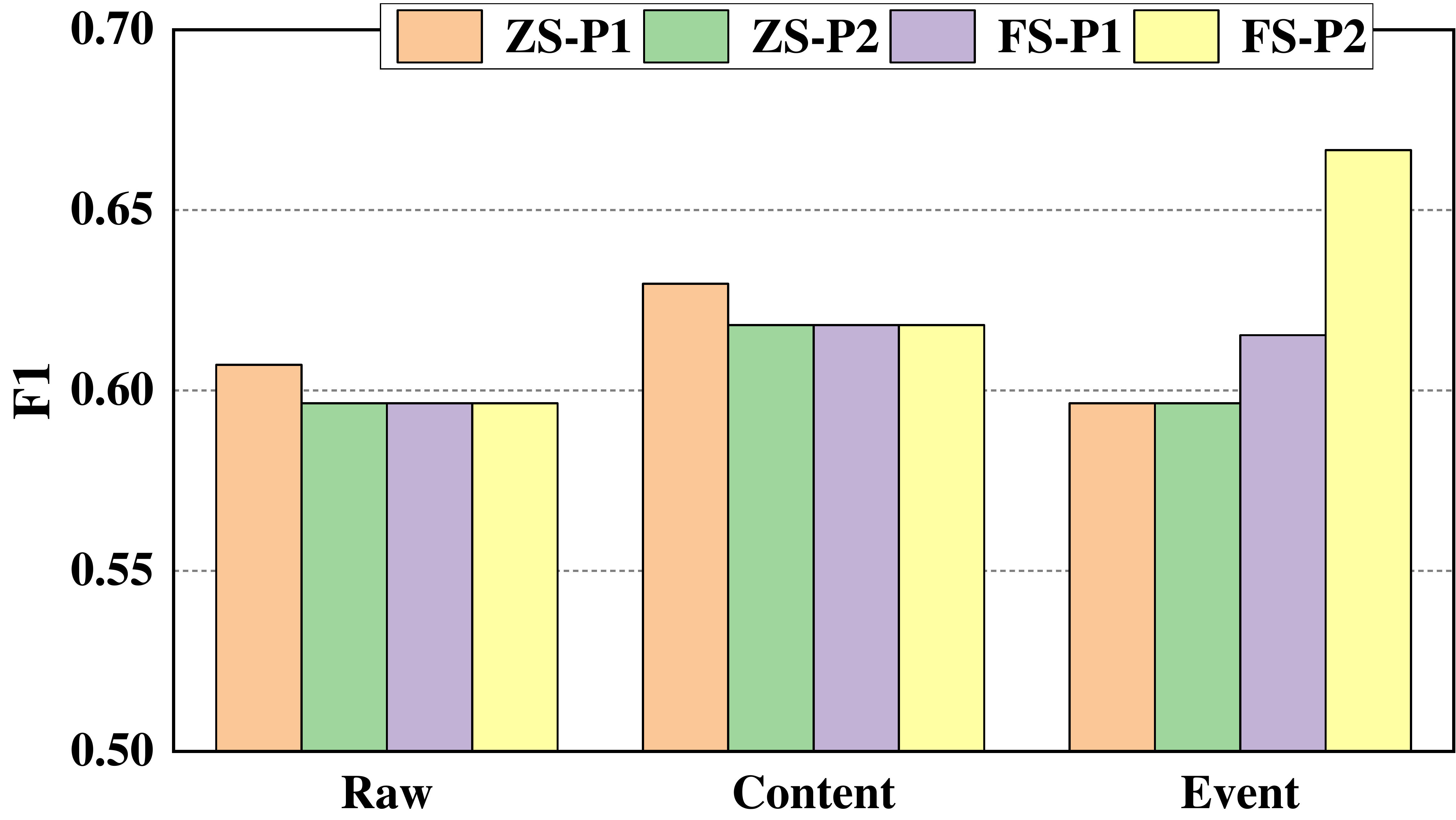} \label{fig:prompt-bgl}} 
    \subfigure[Spirit]{\includegraphics[scale=0.085]{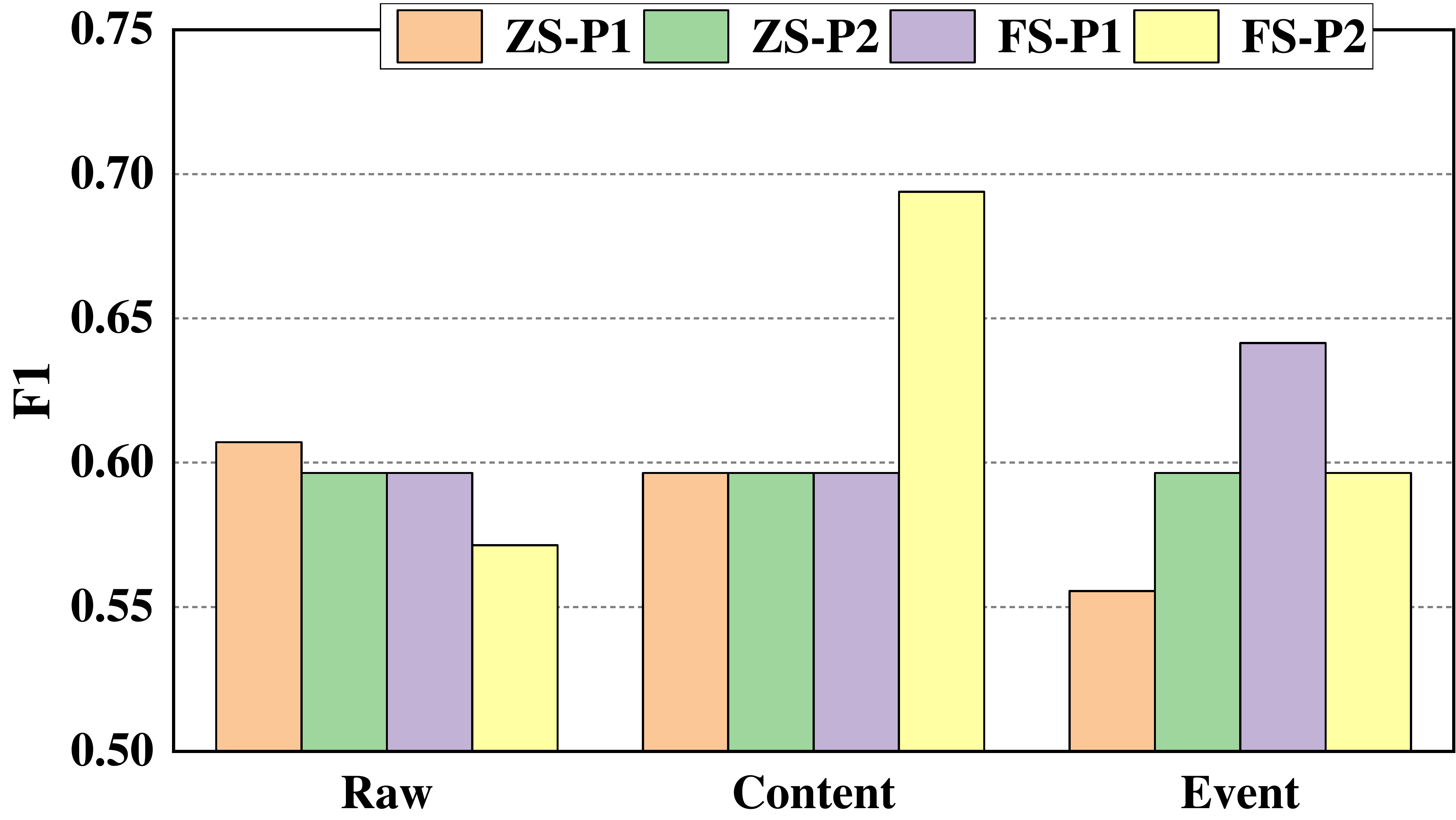} \label{fig:prompt-spirit}} 
    \caption{Performance of LogGPT with different prompts.}
    \label{fig:prompt_impace}
\end{figure*}

\textbf{Experiment Settings.} We investigated the performance of LogGPT with different prompts. Specifically, we used both prompt-1 (P1) and prompt-2 (P2) as shown in Figure \ref{fig:prompt_template}, and evaluated the performance of the three types of input sequences in both zero-shot (ZS) and few-shot (FS) settings, with a fixed window size of 50. The experimental results are presented in Figure \ref{fig:prompt_impace}. 
\par
We can see that for the raw sequence, P1 has a higher F1 score than P2. In particular, it can be seen that human knowledge injection is not always beneficial, since the ZS-P1 with the raw sequence is the highest in terms of F1 score on both datasets. P2 has more advantages than P1 for content sequence and event sequence. For example, FS-P2 has an F1 score of 0.694 on Spirit, which is much higher than other types of inputs. Comparing the three different sequence types, the content sequence performs better than the other two sequence types. We believe this is because the content removes irrelevant information (such as timestamps), which allows ChatGPT to understand its semantic information more precisely. 
\par
\textbf{Summary.}  The prompts have a significant impact on the log-based anomaly detection task. In contrast to expectations, the incorporation of human knowledge does not always result in improved detection performance. It is essential to explore various prompts that align with the specific context in order to attain satisfactory detection performance.

\begin{table}[!htbp]
\centering
\caption{Performance with different injection types on the Spirit dataset}
\label{tab:injection}
\begin{tabular}{cccccc} 
\toprule
\begin{tabular}[c]{@{}c@{}}\textbf{Window}\\\textbf{~Size}\end{tabular} & \begin{tabular}[c]{@{}c@{}}\textbf{Injection }\\\textbf{Type}\end{tabular} & \textbf{F1} & \textbf{P} & \textbf{R} & \textbf{S}  \\ 
\hline
10                                                                      & \multirow{5}{*}{Abnormal}                                                  & 0.026       & 0.100      & 0.015      & 0.932       \\
20                                                                      &                                                                            & 0.000       & 0.000      & 0.000      & 0.000       \\
30                                                                      &                                                                            & 0.490       & 0.545      & 0.444      & \textbf{0.750}       \\
40                                                                      &                                                                            & 0.000       & 0.000      & 0.000      & 0.000       \\
50                                                                      &                                                                            & 0.485       & 0.500      & 0.471      & 0.652       \\ 
\hline\hline
10                                                                      & \multirow{5}{*}{Normal}                                                    & 0.601       & 0.432      & 0.985      & 0.333       \\
20                                                                      &                                                                            & 0.629       & 0.459      & \textbf{1.000}      & 0.246       \\
30                                                                      &                                                                            & \textbf{0.740}       & \textbf{0.587}      & \textbf{1.000}     & 0.525       \\
40                                                                      &                                                                            & 0.714       & 0.556      & \textbf{1.000}      & 0.467       \\
50                                                                      &                                                                            & 0.694       & 0.531      & \textbf{1.000}      & 0.348       \\
\bottomrule
\end{tabular}
\end{table}

\subsubsection{\textbf{RQ2: Performance on different window sizes}}  
\
\par
\textbf{Experiment Settings.} We evaluated the performance of LogGPT with the window size increasing from 10 to 50. P2 is the default prompt. The experimental results are presented in Figure \ref{fig:window} and more results are presented in Appendix \ref{sec:appendix}
\par
We can observe a positive correlation between the F1 score and window size in the majority of cases, indicating that larger window sizes generally result in higher F1 scores. This suggests that incorporating more contextual information is often beneficial. Furthermore, the three types of sequences can be ranked in terms of performance from best to worst as follows: content sequence, event sequence, and raw sequence. This ranking aligns with our expectations, as the raw sequence contains a significant amount of irrelevant information, while the event sequence lacks parameter information. Finally, the results also demonstrate the advantages of the few-shot setting in most cases.
\par
\textbf{Summary.} The window size affects the performance of LogGPT where increasing the window size usually results in higher accuracy in anomaly detection. In general, better performance could be achieved through the content sequence and human knowledge injection.

\begin{figure*}[!tb]
    \centering
    \subfigure[BGL (ZS-P2)]{\includegraphics[scale=0.075]{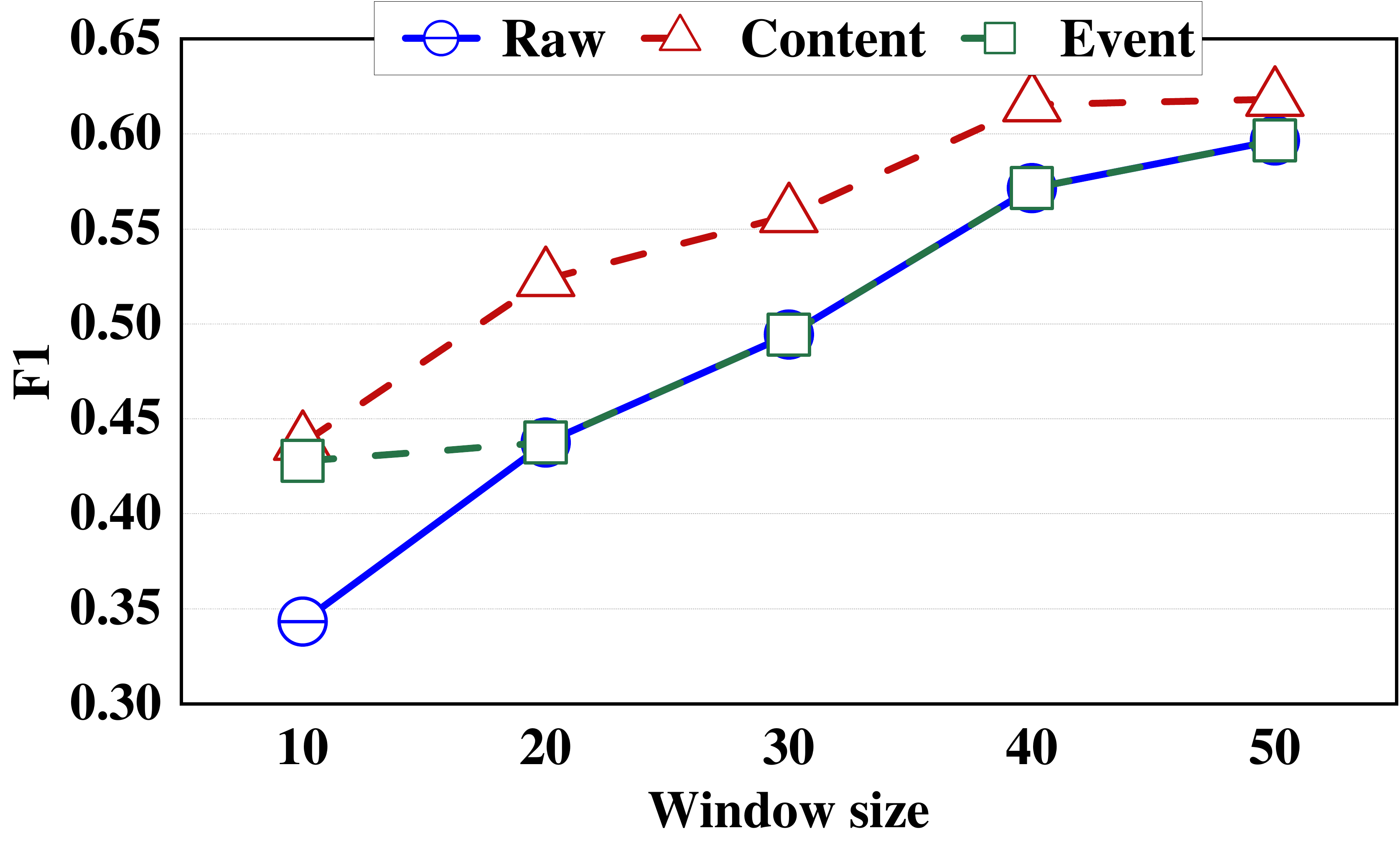} \label{fig:BGL_zeroshot}} 
    \subfigure[BGL (FS-P2)]{\includegraphics[scale=0.075]{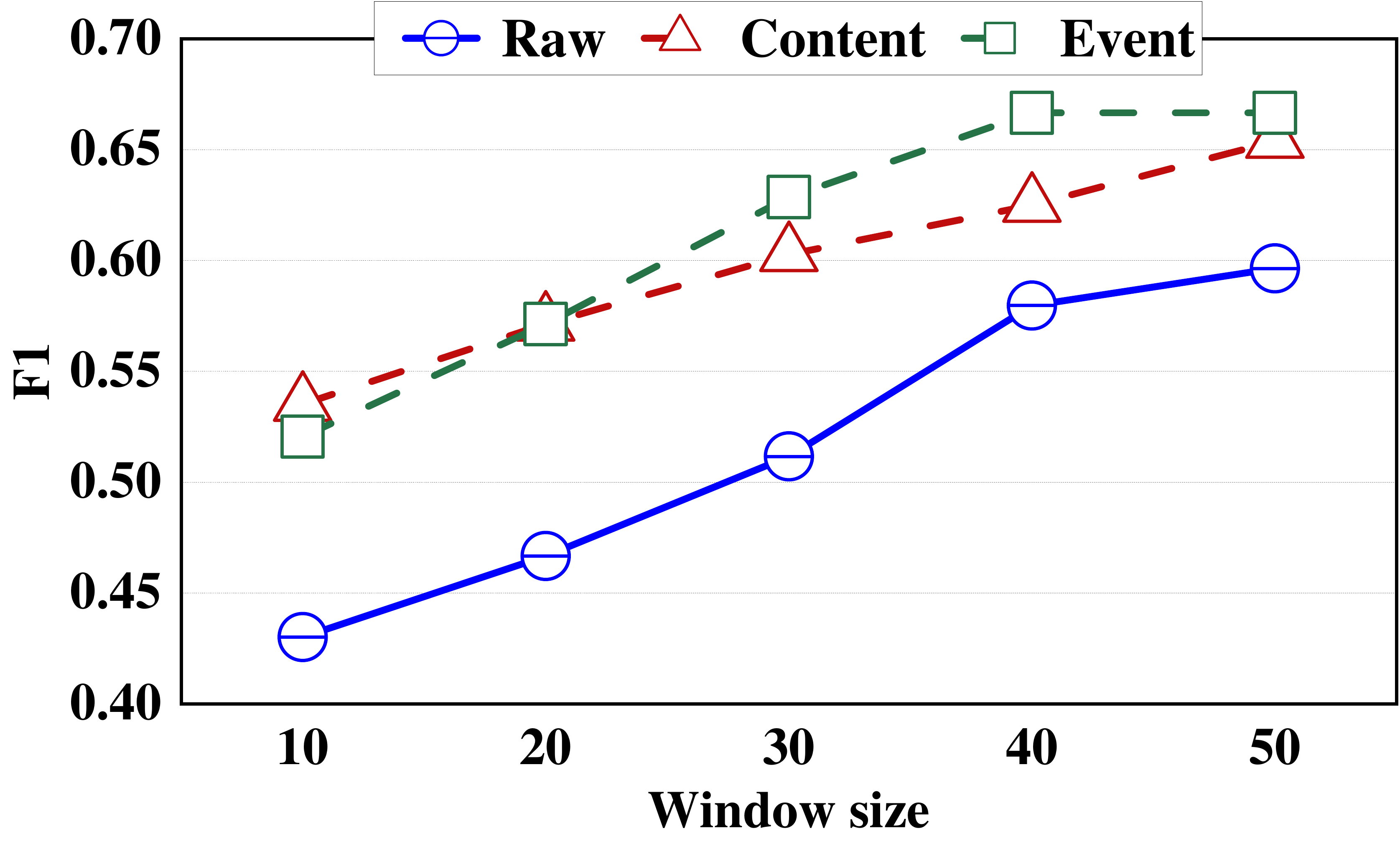} \label{fig:BGL_fewshot}} 
    \subfigure[Spirit (ZS-P2)]{\includegraphics[scale=0.075]{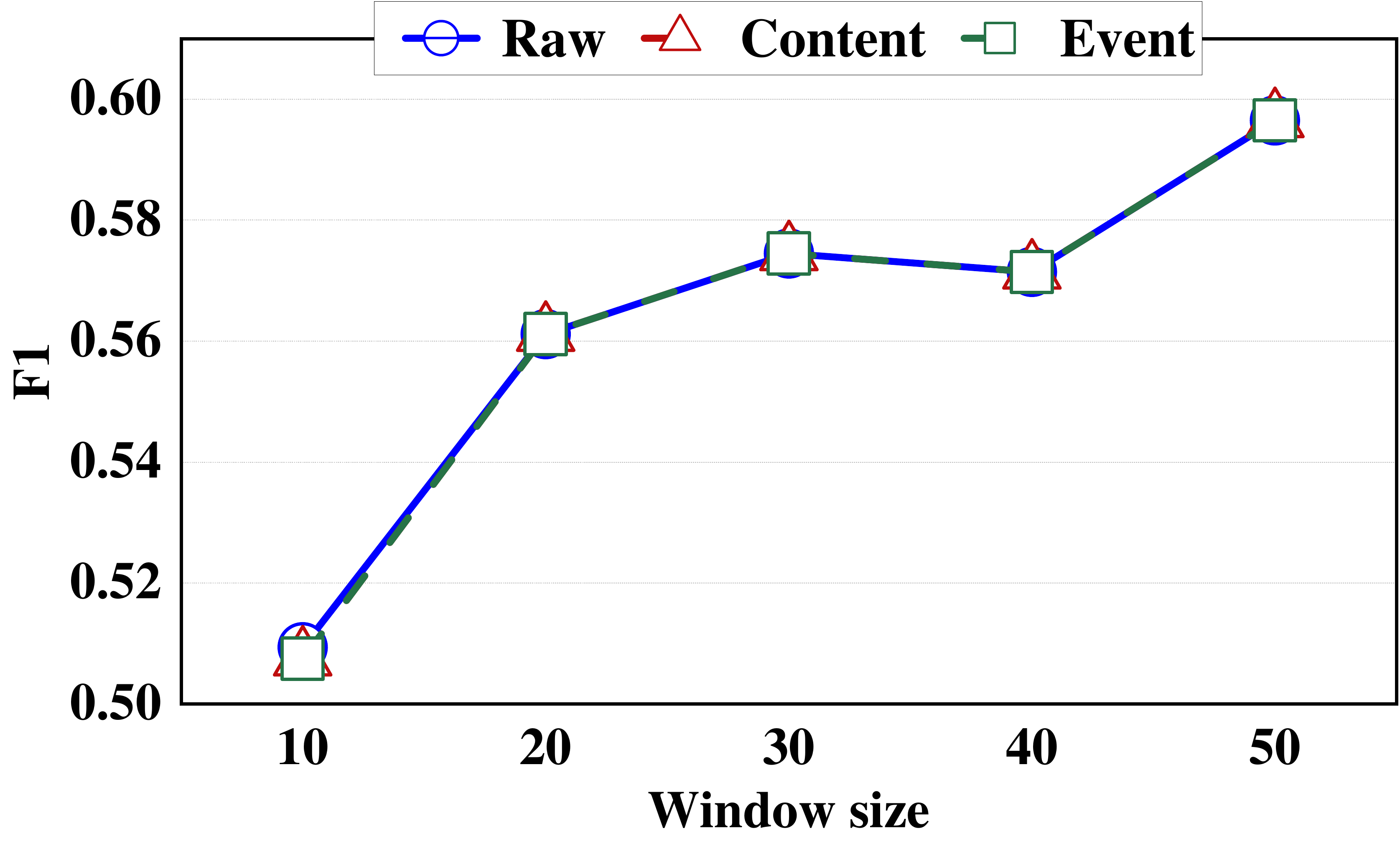} \label{fig:Spirit_zeroshot}} 
    \subfigure[Spirit (FS-P2)]{\includegraphics[scale=0.075]{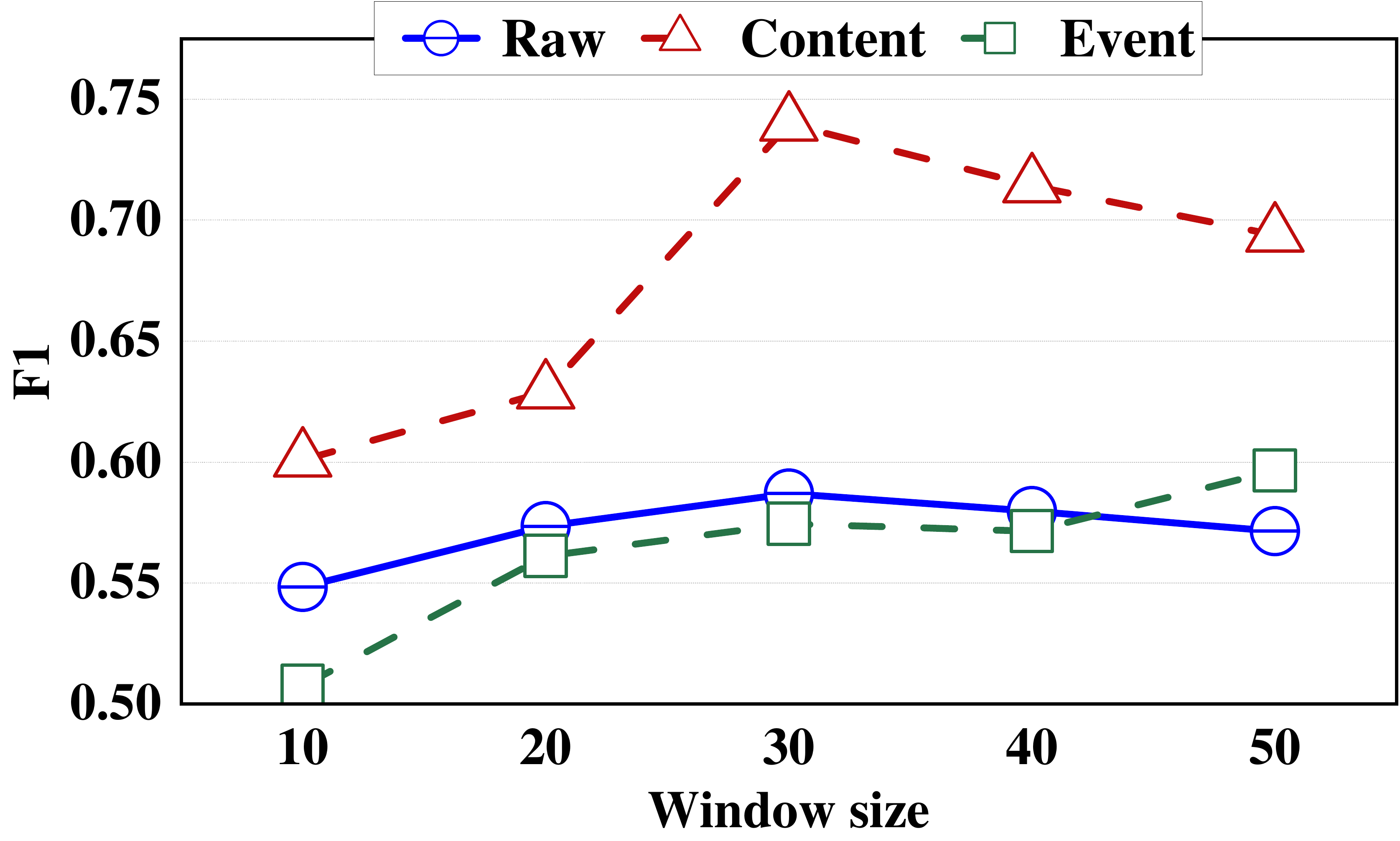} \label{fig:Spirit_fewshot}} 
    \caption{Performance of LogGPT w.r.t different window size.}
    \label{fig:window}
\end{figure*}

\begin{table*}[!hbtp]
\centering
\caption{Performance comparison on BGL and Spirit (Prompt-2 with Content sequence)}
\label{tab:compare}
\resizebox{1.0\linewidth}{!}{
\begin{tabular}{ccccccc||ccccc} 
\toprule
\multirow{2}{*}{\begin{tabular}[c]{@{}c@{}}\textbf{Window }\\\textbf{Size }\end{tabular}} & \multirow{2}{*}{\textbf{Metrics }} & \multicolumn{4}{c}{\textbf{BGL}}    & \multicolumn{6}{c}{\textbf{Spirit}}                                                                    \\ 
\cline{3-12}
&   & \textbf{DeepLog} & \textbf{LogAnomaly} & \textbf{LogRobust} & \begin{tabular}[c]{@{}c@{}}\textbf{LogGPT}\\\textbf{(zero-shot)}\end{tabular} & \begin{tabular}[c]{@{}c@{}}\textbf{LogGPT}\\\textbf{(few-shot)}\end{tabular} & \textbf{DeepLog} & \textbf{LogAnomaly} & \textbf{LogRobust} & \begin{tabular}[c]{@{}c@{}}\textbf{LogGPT}\\\textbf{(zero-shot)}\end{tabular} & \begin{tabular}[c]{@{}c@{}}\textbf{LogGPT}\\\textbf{(few-shot)}\end{tabular}  \\ 
\hline
\multirow{4}{*}{10}   
& F    & 0.168   & 0.201     & \textbf{0.944}   & 0.437  & 0.444  & 0.521     & \textbf{0.679}     & 0.667   & 0.507   & 0.601      \\
& P    & 0.920   & 0.112     & \textbf{0.975}   & 0.280  & 0.286  & 0.352     & 0.602     & \textbf{1.000}   & 0.340   & 0.432      \\
& R    & 0.992   & 0.985     & 0.914   & \textbf{1.000}  & \textbf{1.000}  & \textbf{1.000}     & 0.779     & 0.500   & \textbf{1.000}   & 0.985      \\
& S    & 0.370   & 0.501     & \textbf{0.999}   & 0.356  & 0.375  & 0.053     & 0.735     & \textbf{1.000}   & 0.000   & 0.333      \\ 
\hline
\multirow{4}{*}{20}                                                                       
& F    & 0.219   & 0.228     & \textbf{0.601}   & 0.523  & 0.523  & 0.627     & 0.614     & 0.700   & 0.561   & \textbf{0.629}     \\
& P    & 0.123   & 0.129     & \textbf{0.442}   & 0.354  & 0.354  & 0.591     & 0.500     & \textbf{1.000 }  & 0.390   & 0.459     \\
& R    & 0.977   & 0.975     & 0.938   & \textbf{1.000}  & \textbf{1.000}  & 0.667     & 0.795     & 0.538   & \textbf{1.000}   & \textbf{1.000}      \\
& S    & 0.513   & 0.552     & \textbf{0.919}   & 0.292  & 0.292  & 0.705     & 0.492     & \textbf{1.000}   & 0.000   & 0.246        \\ 
\hline
\multirow{4}{*}{30}                                                                      
& F    & 0.129   & 0.146     & \textbf{0.891}   & 0.557  & 0.571  & 0.581     & 0.632     & \textbf{0.744}   & 0.574   & 0.740    \\
& P    & 0.688   & 0.788     & \textbf{0.893}   & 0.386  & 0.400  & 0.409     & 0.600     & \textbf{1.000}   & 0.403   & 0.587     \\
& R    & \textbf{1.000}   & 0.972     & 0.888   & \textbf{1.000}  & \textbf{1.000}  & \textbf{1.000}     & 0.667     & 0.593   & \textbf{1.000}   & \textbf{1.000}      \\
& S    & 0.082   & 0.180     & \textbf{0.992}   & 0.222  & 0.267  & 0.025     & 0.700     & \textbf{1.000}   & 0.000   & 0.525      \\ 
\hline
\multirow{4}{*}{40}                                                                      
& F    & 0.147    & 0.239    & 0.374   & 0.615  &  \textbf{0.625}  & 0.597     & 0.711       & 0.667      & 0.571     & \textbf{0.714}      \\
& P    & \textbf{0.793}    & 0.137    & 0.236   & 0.444  &  0.455  & 0.426     & 0.640       & \textbf{1.000}      & 0.400     & 0.556       \\
& R    & 0.972    & 0.960    & 0.900   & \textbf{1.000} &  \textbf{1.000}  & \textbf{1.000}     & 0.800       & 0.500      & \textbf{1.000}     & \textbf{1.000}      \\
& S    & 0.137    & 0.541    & \textbf{0.780}   & 0.167  &  0.200  & 0.100     & 0.700       & \textbf{1.000}      & 0.000     & 0.467   \\ 
\hline
\multirow{4}{*}{50}                                                                       
& F    & 0.224   & 0.243     & 0.304   & \textbf{0.618}  &  \textbf{0.618}   & 0.607      & 0.571               & 0.455              & 0.596   & \textbf{0.694}     \\
& P    & 0.126   & 0.139     & 0.183   & \textbf{0.447}  &  \textbf{0.447}   & 0.436      & 0.410               & \textbf{1.000}              & 0.425   & 0.531       \\
& R    & 0.994   & 0.942     & 0.911   & \textbf{1.000}  &  \textbf{1.000}   & \textbf{1.000}      & 0.941               & 0.294              & \textbf{1.000}   & \textbf{1.000}       \\
& S    & 0.418   & 0.537     & \textbf{0.676}   & 0.087  &  0.087   & 0.043      & 0.000               & \textbf{1.000}              & 0.000    & 0.348       \\
\bottomrule
\end{tabular}}
\end{table*}

\subsubsection{\textbf{RQ3: The impact of human knowledge injection}}
\
\par
\textbf{Experiment Settings.} We studied the effects of different human knowledge injections. Two types of knowledge injection (abnormal and normal) are employed, we randomly sample 5 normal logs and 5 abnormal logs from the Spirit dataset. The experimental results are presented in Table \ref{tab:injection}.
\par
We can observe that the performance of LogGPT varies a lot with different types of knowledge injection. For example, abnormal knowledge injection results in an F1 score range of $\left[0.000, 0.490\right]$ with the window size increasing from 10 to 50. When experimenting with the normal knowledge injection, it achieves a promising performance with the F1 scores ranging from 0.601 to 0.740. Because the abnormal type is diverse, a few abnormal samples cannot cover most types. The results also show that the recall remains high across all window sizes, while precision and specificity are low. This suggests that LogGPT frequently misses identical normal sequences as abnormal, resulting in a high false positive rate. 
\par
\textbf{Summary.} Different human knowledge injections have a significant impact on the performance of log-based anomaly detection. LogGPT adopts a conservative approach by identifying only highly certain logs as normal and identifying the remaining logs as abnormal. This bias can be attributed to the presence of many low-level alert events in the logs, which are often disregarded by administrators despite being semantically categorized as abnormal.
\begin{table*}[!htbp]
\centering
\caption{Interpretability of detected anomalies on BGL and Spirit}
\label{tab:interpretability}
\renewcommand\arraystretch{1.2}
\begin{tabular}{ccc} 
\toprule
\textbf{Dataset}     & \textbf{Key}                                                    & \textbf{Response}                                                                                                                                                           \\ 
\hline
\multirow{11}{*}{BGL} & \begin{tabular}[c]{@{}c@{}}Anomaly~~\\log sequence\end{tabular} & \begin{tabular}[c]{@{}c@{}}… \\RAS KERNEL \textcolor{red}{FATAL data TLB error interrupt}\\…\end{tabular}                                                                                    \\
                     & Reports       & data TLB error interrupt~ ~ ~~                                                                                                                                              \\
                     & \begin{tabular}[c]{@{}c@{}}Preventive~\\measures\end{tabular}   & \begin{tabular}[c]{@{}c@{}}Check the memory and CPU usage\\Check the system logs~ ~ ~~\end{tabular}                                                                           \\
                     \cline{2-3}
                     & \begin{tabular}[c]{@{}c@{}}Anomaly~~\\log sequence\end{tabular} & \begin{tabular}[c]{@{}c@{}}\textbf{…}\\RAS KERNEL INFO \textcolor{red}{MACHINE CHECK DCR read timeout}\\\textbf{…}\end{tabular}  \\
                     & Reports                                                         & RAS
KERNEL FATAL Lustre mount FAILED : bglio78 : block\_id : location~ ~ ~~                                                                                                 \\
                     & \begin{tabular}[c]{@{}c@{}}Preventive~\\measures\end{tabular}   & Check the Lustre mount FAILED~ ~ ~~                                                                                                                                         \\
\hline\hline    
\multirow{11}{*}{Spirit} & \begin{tabular}[c]{@{}c@{}}Anomaly~~\\log sequence\end{tabular} & \begin{tabular}[c]{@{}c@{}}… \\ \textcolor{red}{pbs\_mom: Connection refused (111) in open\_demux,} \\ open\_demux: connect 172.30.71.32:44020\\…\end{tabular}                                                                                    \\
                     & Reports                                                         & Anomaly detected: Multiple sshd sessions opened for user root ~ ~ ~~                                                                                                                                              \\
                     & \begin{tabular}[c]{@{}c@{}}Preventive~\\measures\end{tabular}   & \begin{tabular}[c]{@{}c@{}}Limit the number of sshd sessions opened for user root~ ~ ~~\end{tabular}                                                                           \\
                     \cline{2-3}
                     & \begin{tabular}[c]{@{}c@{}}Anomaly~~\\log sequence\end{tabular} & \begin{tabular}[c]{@{}c@{}}\textbf{…}\\ \textcolor{red}{pbs\_mom: Connection refused (111) in open\_demux}, \\ open\_demux: connect 172.30.71.32:43908  \\\textbf{…}\end{tabular}  \\
                     & Reports                                                         & Inappropriate ioctl for device (25) in search\_env\_and\_open, failed connect to mpiexec process on MS~ ~ ~~                                                                                                 \\
                     & \begin{tabular}[c]{@{}c@{}}Preventive~\\measures\end{tabular}   & Check the mpiexec process on MS~ ~ ~~                                                                                                                                         \\
\bottomrule
\end{tabular}
\end{table*}

\subsubsection{\textbf{RQ4: Performance comparison to three baseline methods}}  
\
\par
\textbf{Experiment Settings.} We evaluated the performance of LogGPT with both zero-shot setting and few-shot setting and used prompt-2 as the default prompt. The experimental results are shown in Table \ref{tab:compare}.
\par
Compared to the three deep learning-based methods, LogGPT with few-shot setting achieved superior performance on both datasets in terms of the F1 score, precision, and recall. And LogGPT with the zero-shot setting achieved a competitive F1 score on both datasets. An interesting observation is that LogGPT achieved the same F1 score between zero-shot and few-shot settings on BGL when the window size is 20 and 50, suggesting that adding some prior knowledge is not always beneficial. Furthermore, LogGPT achieves the highest recall, but the specificity is underperforming, especially on Spirit where the zero-shot setting is 0. This demonstrated that LogGPT is more reserved, resulting in the identification of many normal logs as anomalies. 
\par
\textbf{Summary.} LogGPT shows promising performance on both the BGL and Spirit datasets. Its ability to identify log sequences as normal or anomalous suggests its potential as an effective tool for log anomaly detection in various domains.

\subsubsection{\textbf{RQ5: Interpretability of anomalies by LogGPT}}  
\
\par
Providing more explanations for the anomalies detected by the models has been difficult since DL-based methods are considered a black box whose decision-making process is not explainable, which poses challenges for administrators in timely identifying and preventing anomalies. In this experiment, we investigate the potential of LogGPT for interpretability with respect to anomaly localization and prevention. 
\par
\textbf{Experiment Settings.} As LogGPT(few-shot) achieved the highest recall on BGL and Spirit datasets, we randomly selected two anomalous raw sequences on both datasets, respectively, to verify the effectiveness of anomaly localization and anomaly prevention. 
\par
The results are presented in Tabel \ref{tab:interpretability}. We can see that the response of LogGPT can not only report the cause of the anomaly but also give suggestions for prevention. For example, an anomaly occurred which is \textit{connection refused} during the running of a Linux cluster. LogGPT reports the reason to be the root user opening too many \textit{sshd sessions}. Furthermore, LogGPT suggests to enforce the limitations on the number of \textit{sshd sessions} the root user can open. However, LogGPT's suggestions may not always be effective and they occasionally lead to confusion. For example, in the case of the first example on the BGL dataset, LogGPT suggests that users should first check the CPU and memory usage. If the anomaly persists, LogGPT suggests the administrators to manually analyze historical logs. We can see from these results that while LogGPT is a powerful log-based anomaly detector, its suggestions are generated based on patterns and information extracted from the input sequences. Therefore, there are instances where its suggestions may not be useful or may require additional manual analysis by administrators.
\par
\textbf{Summary.} LogGPT can help administrators find abnormal locations and provide preventive suggestions which improve the efficiency of troubleshooting. However, there are some ineffective outputs in our experiments due to the hallucination problem of ChatGPT. Since this work is not concerned with solving this problem, we only report the existence of ineffective output. We think more specific prompts may be able to reduce such output.

\section{Related works}
In this section, we briefly review the related work of log-based anomaly detection and large language models (LLMs).
\par
\textbf{Log-based anomaly detection} is an important technique for monitoring system activities and identifying suspicious behaviors. Many works have been proposed in recent years \cite{du2017deeplog, zhang2019robust, meng2019loganomaly}, these works can be grouped into three types according to the training strategy.
\par
1) \textbf{Supervised methods} assume that have a large number of labeled data to train binary classification models \cite{zhang2019robust}, which commonly achieved the optimal performance. Zhou et.al \cite{zhou2020logsayer} proposed LogSyaer, which is a log pattern-driven anomaly detection model that addresses these challenges. LogSayer utilizes statistical features, LSTM neural networks, and a BP neural network for adaptive anomaly decisions. Du et al. \cite{du2021log} proposed LogAttention which embeds log patterns into semantic vectors and uses a self-attention-based neural network to detect anomalies. The input to these methods typically is a log event sequence. However, the parameters within the logs also contain valuable information for identifying anomalies. Huang et al. \cite{huang2020hitanomaly} proposed HitAnomaly, which utilizes a hierarchical transformer model to capture both log event sequences and parameters, to improve the detection accuracy. 
\par
2) \textbf{Semi-supervised methods} assume that training takes place only on normal data that is free of anomalies. Its primary idea is to make the models capture the normal sequence patterns, while the abnormal sequence patterns are quite distinct. Some works focus on reconstruction-based methods that utilize generative models, such as AutoEncoders (AEs) \cite{zhang2021logattn} and Generative Adversarial Networks (GANs) \cite{zhao2022trine}, to encode the input log sequences and then attempt to reconstruct the input using a decoder. Any input data that is fed into an already trained model and yields a high reconstruction error is then considered anomalous. Another technical route is using sequential models, such as Recursive Neural Networks (RNNs) \cite{du2017deeplog} and Transformer \cite{devlin2018bert},  to learn normal sequential patterns and predict the next possible log event. If a new log sequence does not match the predicted then it is identified as anomalous. Moreover, we note that many works \cite{zhang2021logattn} claim that it is unsupervised, but it is actually trained using only normal logs.
\par
3) \textbf{Unsupervised methods} \cite{otomo2019latent, bursic2020anomaly} assume that no labels are available for training models. Han et al. \cite{han2021unsupervised} proposed a domain adaptation framework called LogTAD, which makes log data from different systems have similar distributions, enabling the detection model to identify anomalies across multiple systems. Nedelkoski et al. \cite{nedelkoski2020self} focus on learning log representations that capture semantic differences between normal and anomaly logs, improving generalization on unseen logs. It leverages auxiliary log datasets available on the internet to enhance the representation of normal data while providing diversity to prevent overfitting. However, due to the lack of prior knowledge, the detection accuracy is lower than supervised and semi-supervised methods in most scenarios.
\par
\textbf{Large language models (LLMs)} have revolutionized the field of natural language processing (NLP) and have gained significant attention in recent years. OpenAI introduced GPT-1 \cite{radford2018improving}, which demonstrated promising results by pre-training on a large corpus of internet text data and fine-tuning it for specific tasks. After that, most LLMs have been proposed, such as BERT \cite{devlin2018bert}, T5 \cite{raffel2020exploring}, and GPT-3 \cite{brown2020language}, etc. A key milestone in LLM development is InstructGPT \cite{ouyang2022training}, which introduces a framework for instruction fine-tuning of pre-trained language models using Reinforcement Learning from Human Feedback (RLHF). This framework enables LLMs to adapt to a wide range of NLP tasks, enhancing their versatility and flexibility by incorporating human feedback. Unlike LLMs trained alone on text corpora through unsupervised pre-training, RLHF allows models to align with human preferences and values, significantly improving their performance. OpenAI has leveraged similar techniques to develop ChatGPT, a conversation-based language model that brings AI to the forefront, transitioning it from being behind the scenes. Subsequently, OpenAI released a version based on GPT-4 \cite{openai2023gpt4}. Concurrently, other peers have also released similar language models. For example, Meta AI's LLaMA \cite{touvron2023llama} and Google's PaLM \cite{chowdhery2022palm}, etc. 
\par
In terms of applications, LLMs have shown impressive performance in various domains. For example, Microsoft released the Copilot plugin \cite{chen2021evaluating}, a dedicated model for code generation, which greatly improves the efficiency of developers. Additionally, such models as Stable Diffusion, Midjourney, and DALL-E \cite{borji2022generated} have also demonstrated excellent results in artificial intelligence generative content (AIGC). We can envision that LLMs will play a crucial role in numerous decision-making scenarios, paving the way for advancements in various domains. 

\section{Limitations}
Based on our observations, we highlight several limitations of LogGPT:
\begin{itemize}
    \item \textbf{Sensitivity to prompt variation:}  Designing the optimal prompt is not always straightforward, even though prompts play a critical role in LogGPT's performance.
    \item \textbf{Limited window size:} Balancing the need for a larger window size with computational constraints remains a challenge.
    \item \textbf{High false positive rate:}  LogGPT suffers from a high false positive rate, leading to a significant number of incorrect anomaly identifications.
    \item \textbf{Trustworthiness:} The presence of ineffective outputs due to the hallucination problem of ChatGPT.  
\end{itemize}
\par
Addressing these limitations will improve the applicability and effectiveness of LogGPT, and benefit the application of LogGPT in real-world scenarios.

\section{Conclusion and Future Work}
In this study, we proposed LogGPT, a framework for log-based anomaly detection based on ChatGPT. Experiments demonstrated that LogGPT has a promising ability for this important task and better anomalous interpretability, except that many normal sequences are identified anomalies and it causes many false alarms. Moreover, we investigated different prompts construction, window sizes, and the types of input sequence impact on performance. We believe that the results and findings of our study can provide valuable insights into the strengths and limitations of ChatGPT in log-based anomaly detection. As our future work, we will investigate the following:
\begin{itemize}
    \item Experiment with more public log datasets and test more prompts to improve performance.
    \item Try using more large language models such as GPT-4 or LLaMA.
    \item Further exploration of ChatGPT's potential for anomalous interpretability.
\end{itemize}

\bibliographystyle{IEEEtran}
\bibliography{ref}

\section{Appendix} \label{sec:appendix}
\begin{figure*}[!htb]
    \centering
    \subfigure{\includegraphics[scale=0.055]{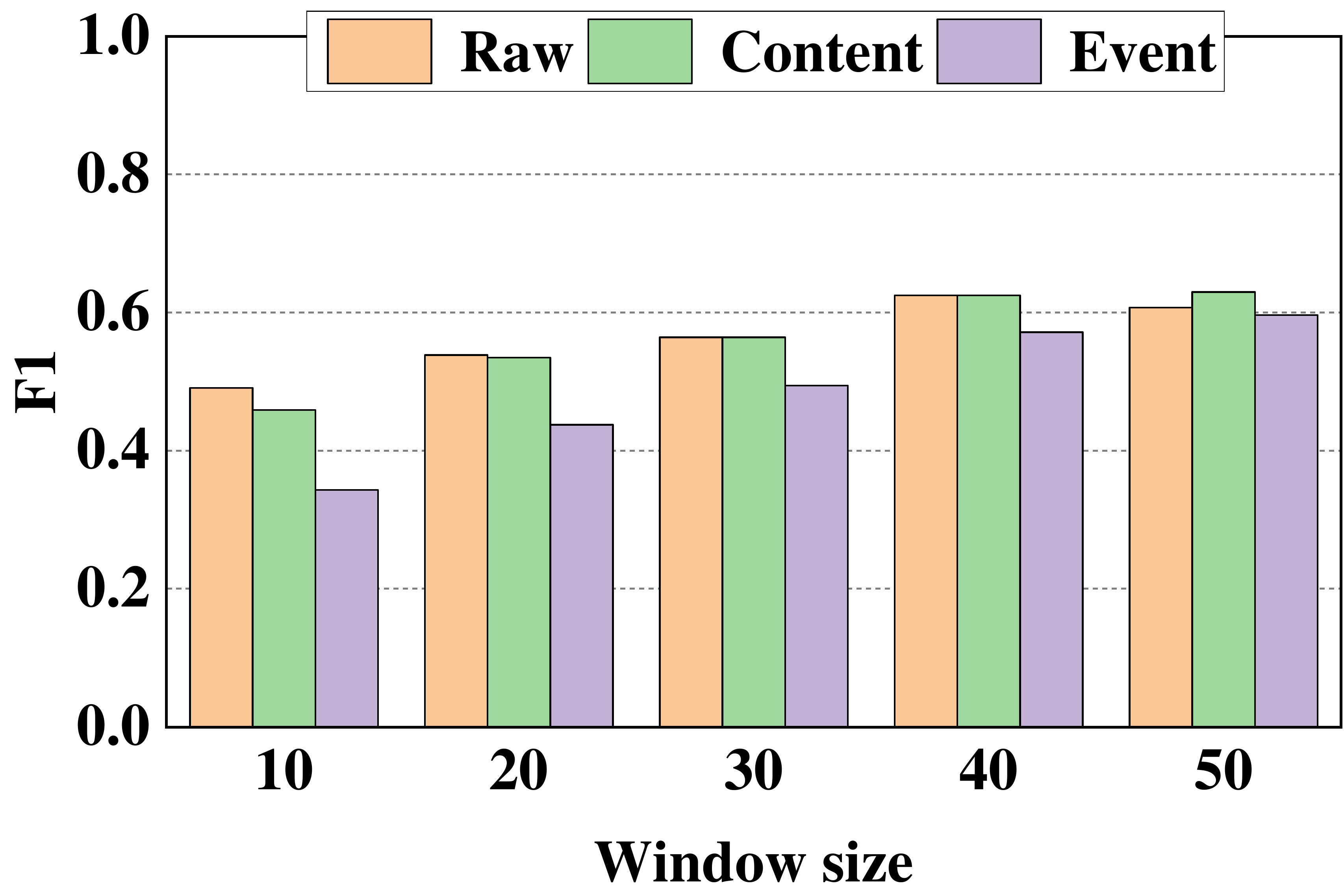}}
    \subfigure{\includegraphics[scale=0.055]{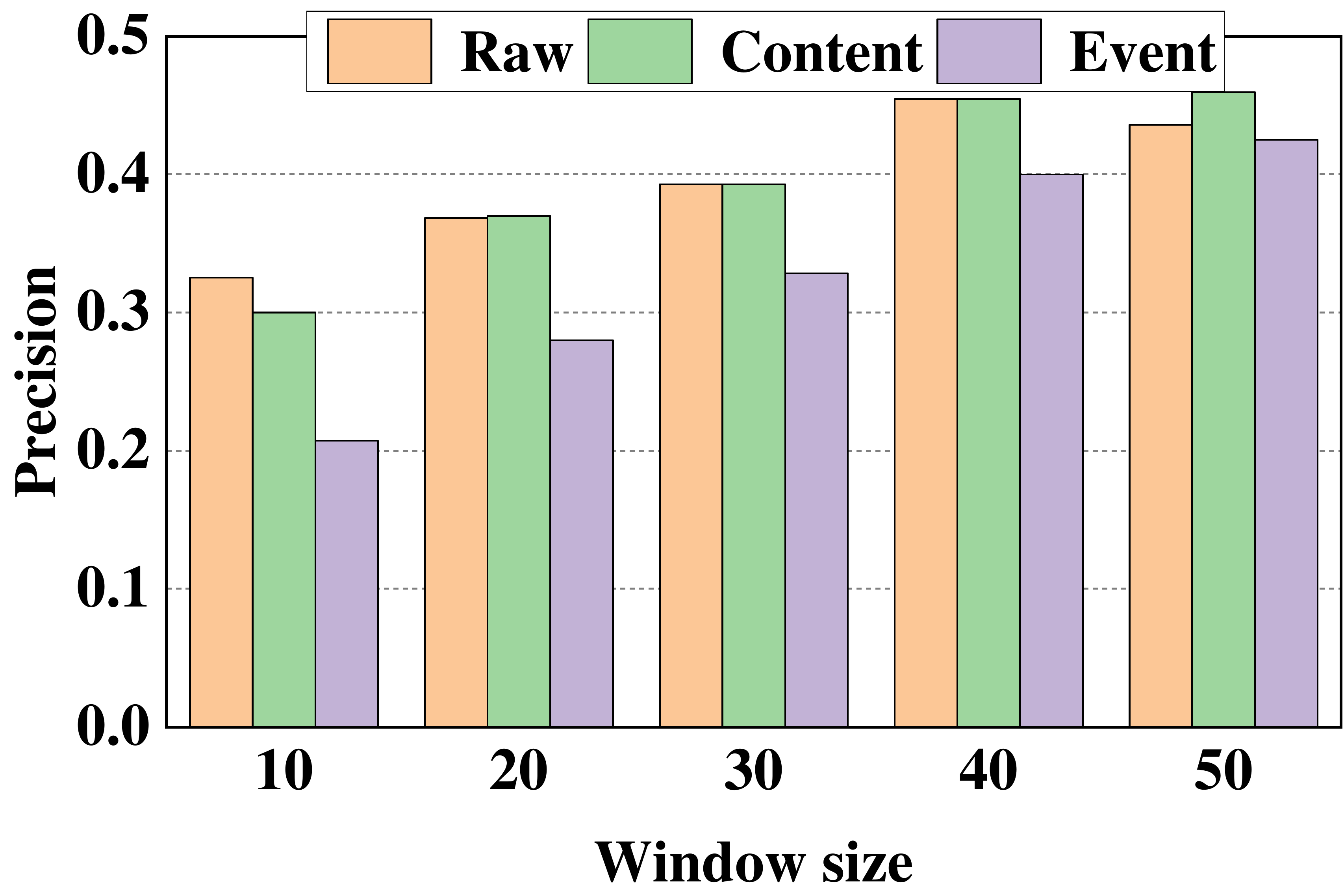}}
    \subfigure{\includegraphics[scale=0.055]{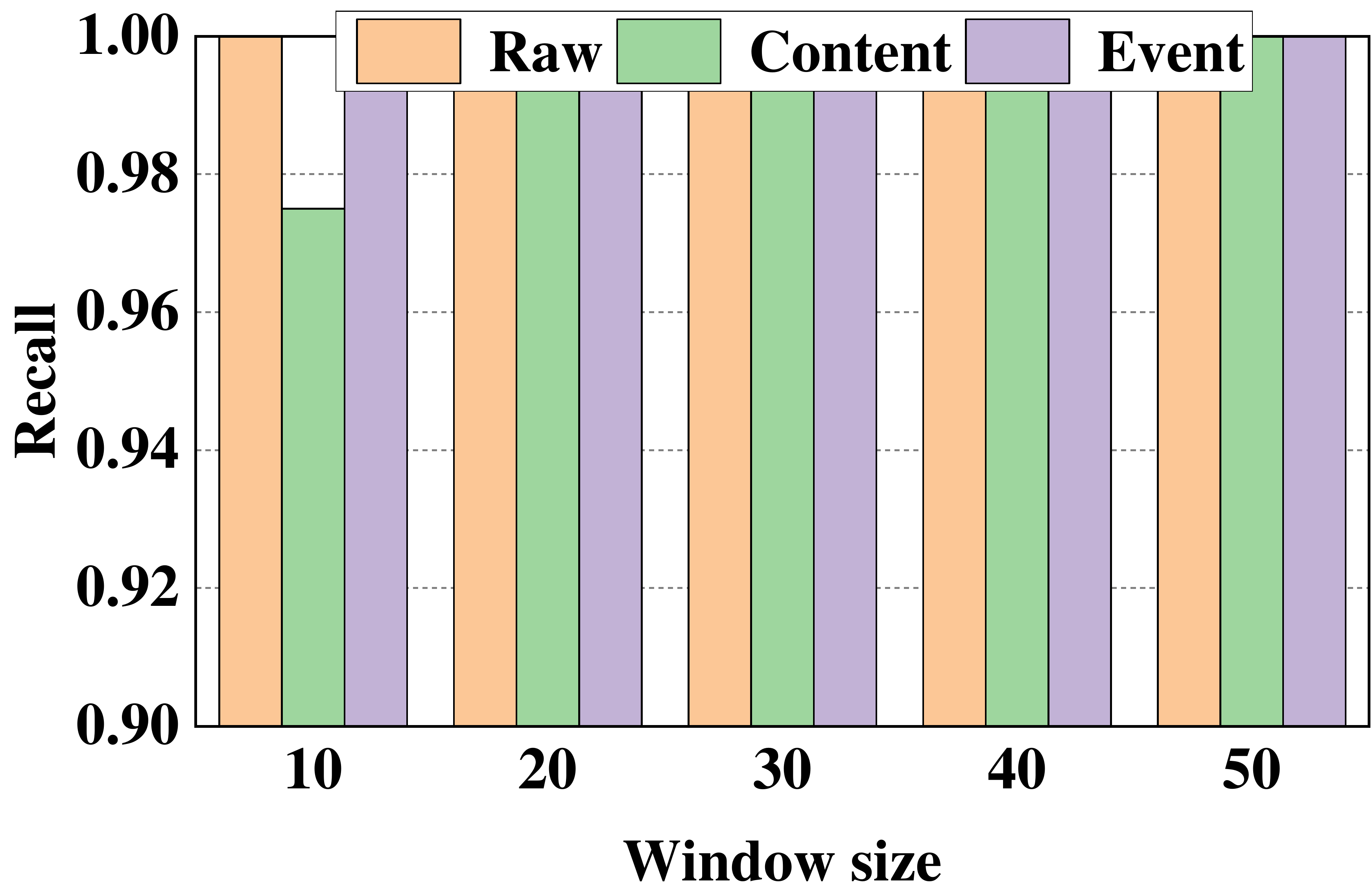}}
    \subfigure{\includegraphics[scale=0.055]{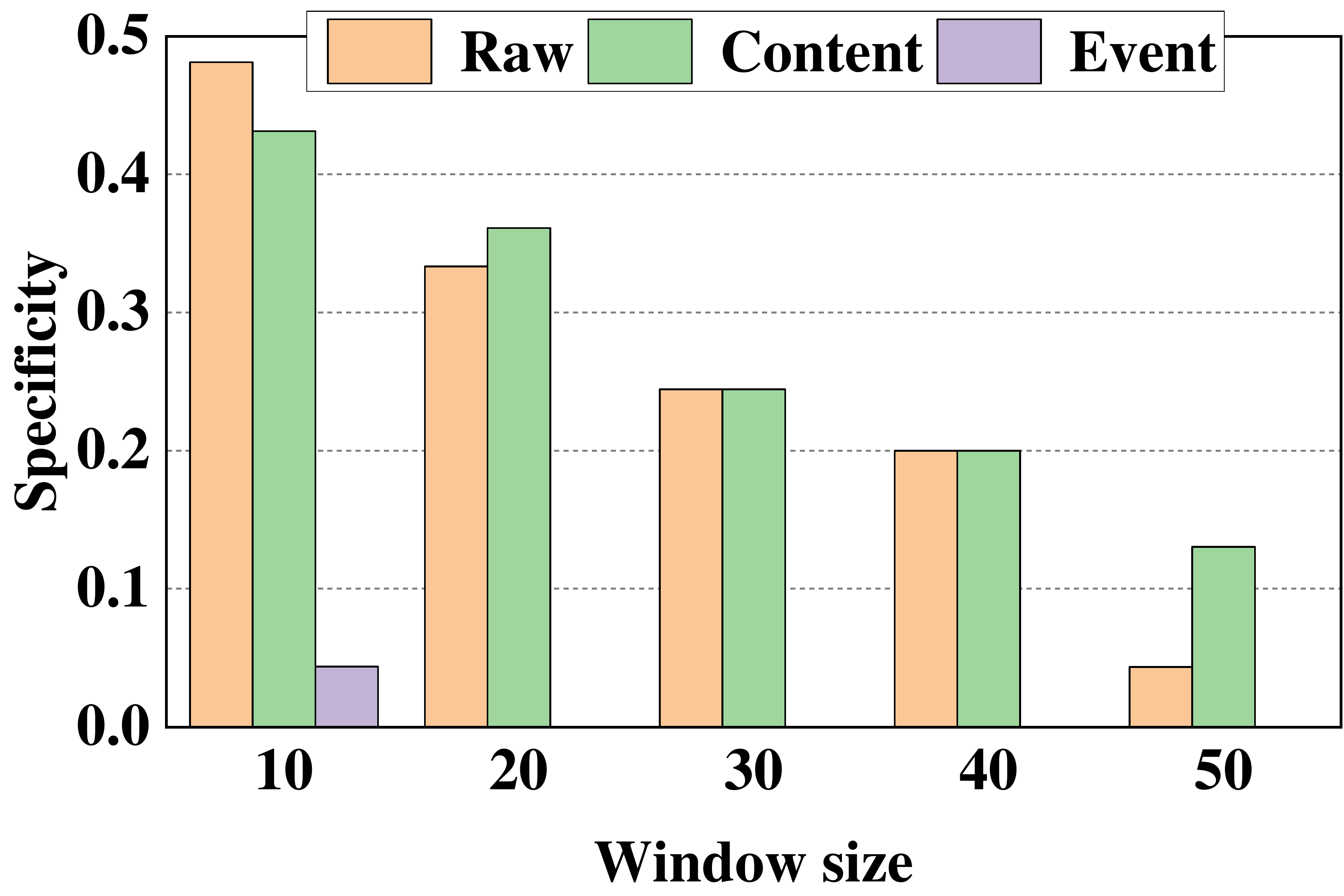}}
    \\
    \subfigure{\includegraphics[scale=0.055]{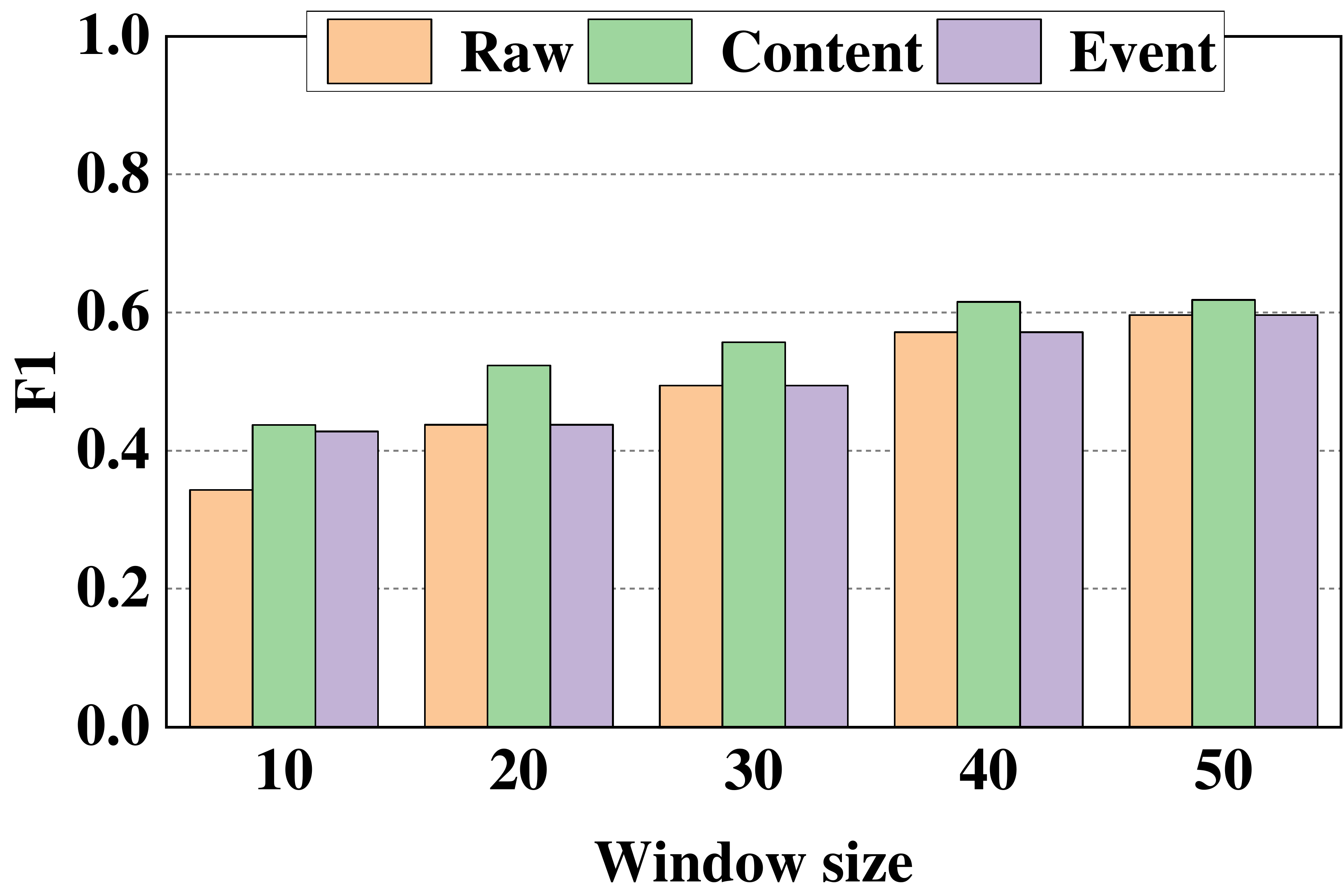}}
    \subfigure{\includegraphics[scale=0.055]{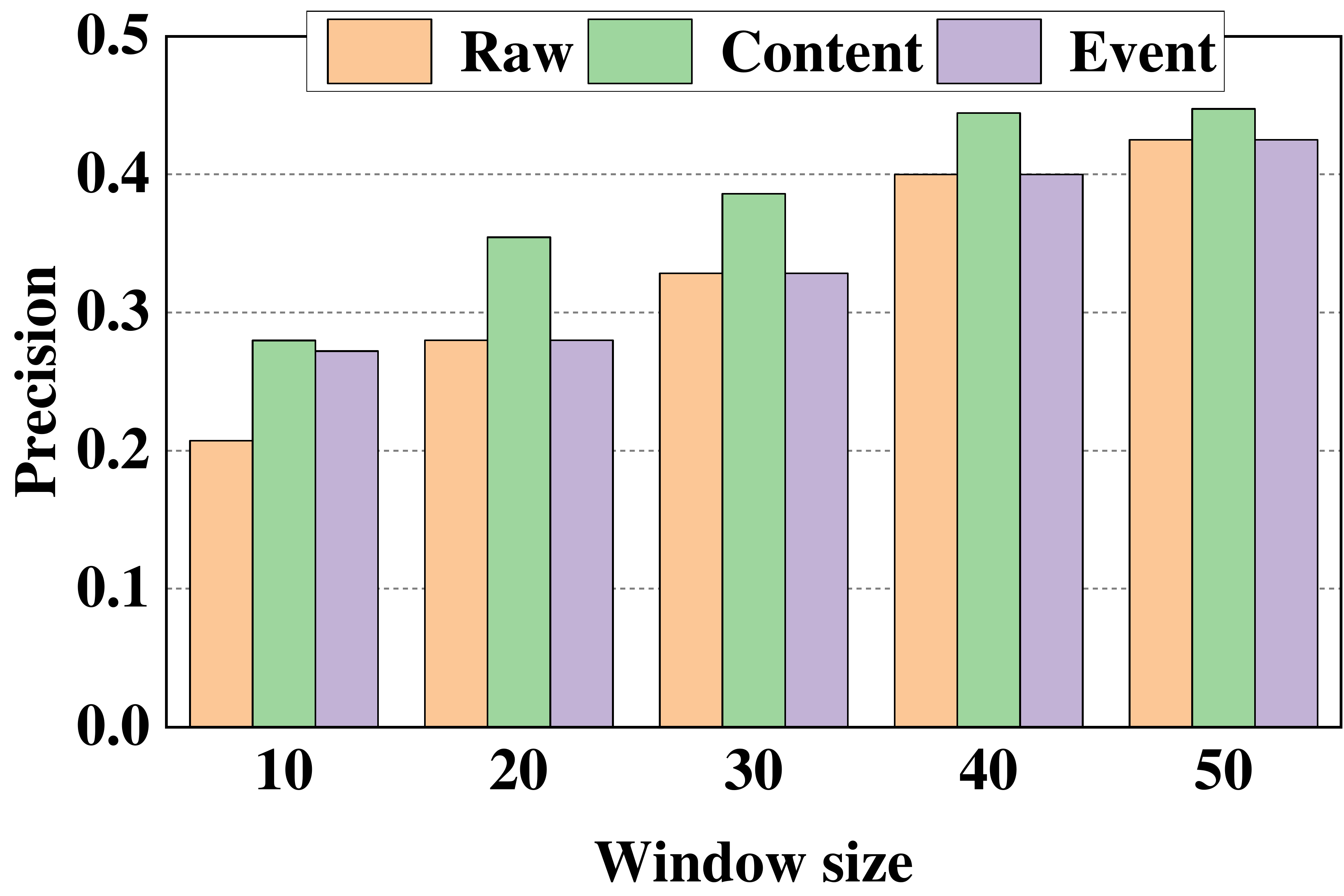}}
    \subfigure{\includegraphics[scale=0.055]{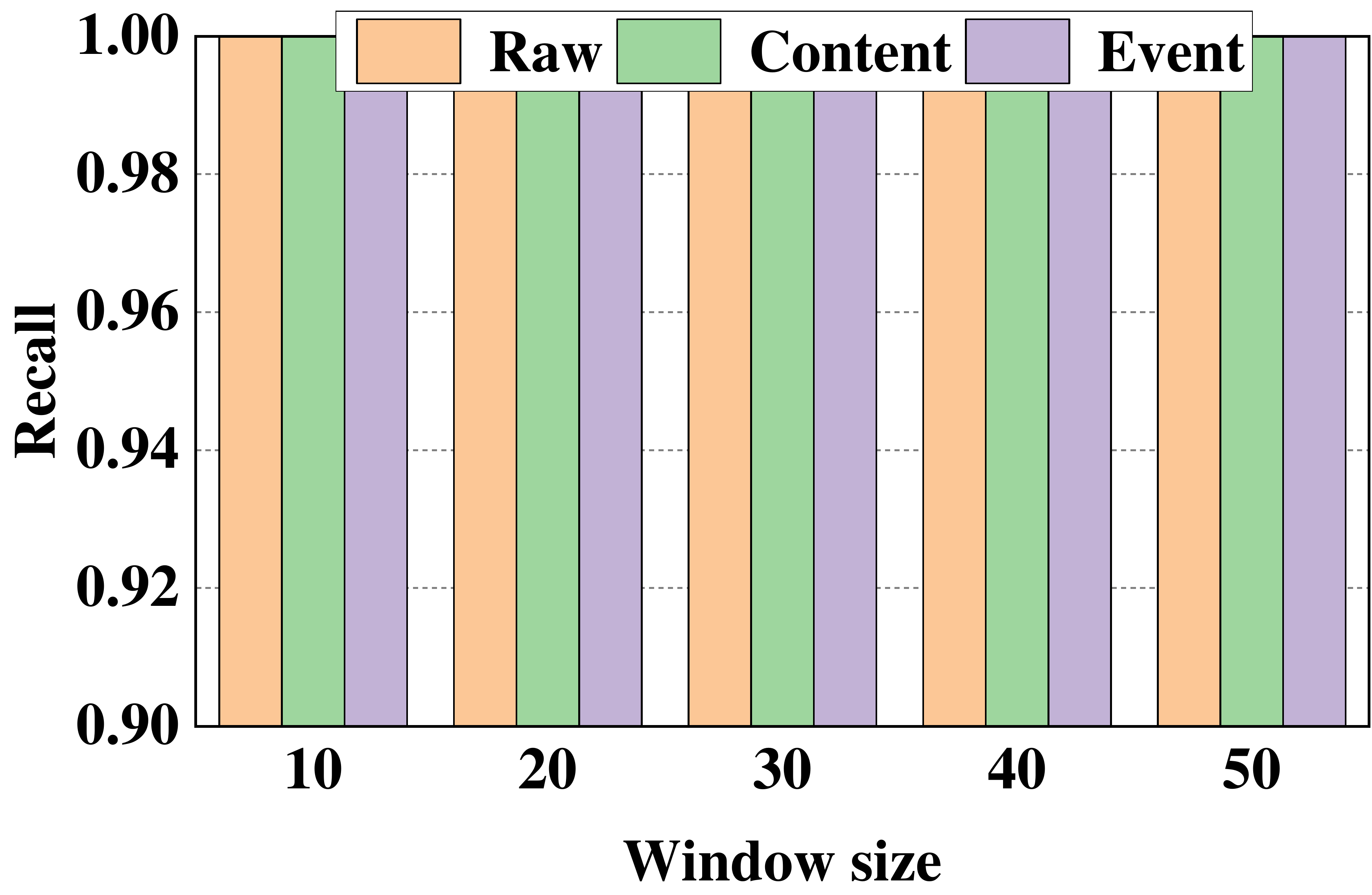}}
    \subfigure{\includegraphics[scale=0.055]{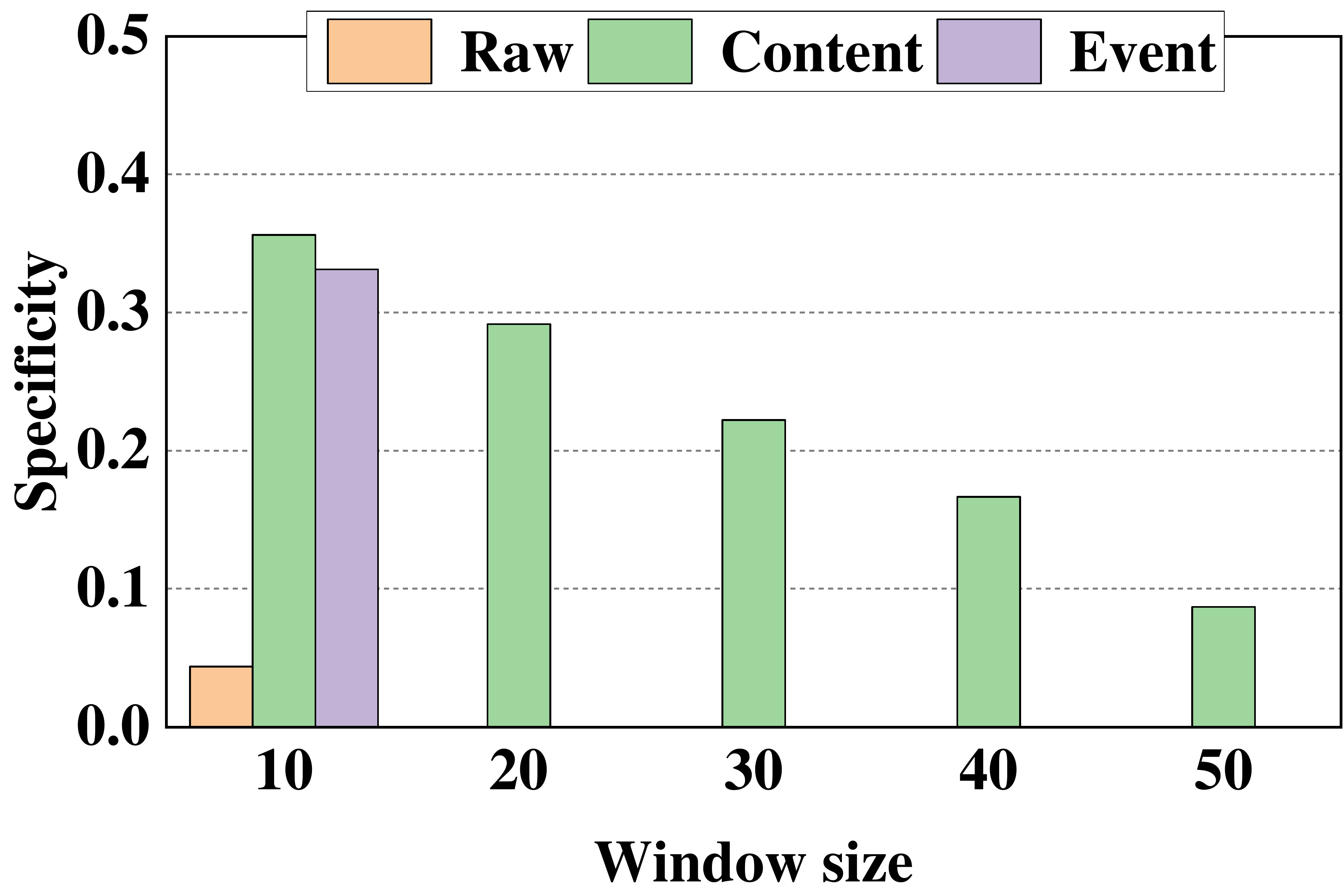}}
    \\
    \subfigure{\includegraphics[scale=0.055]{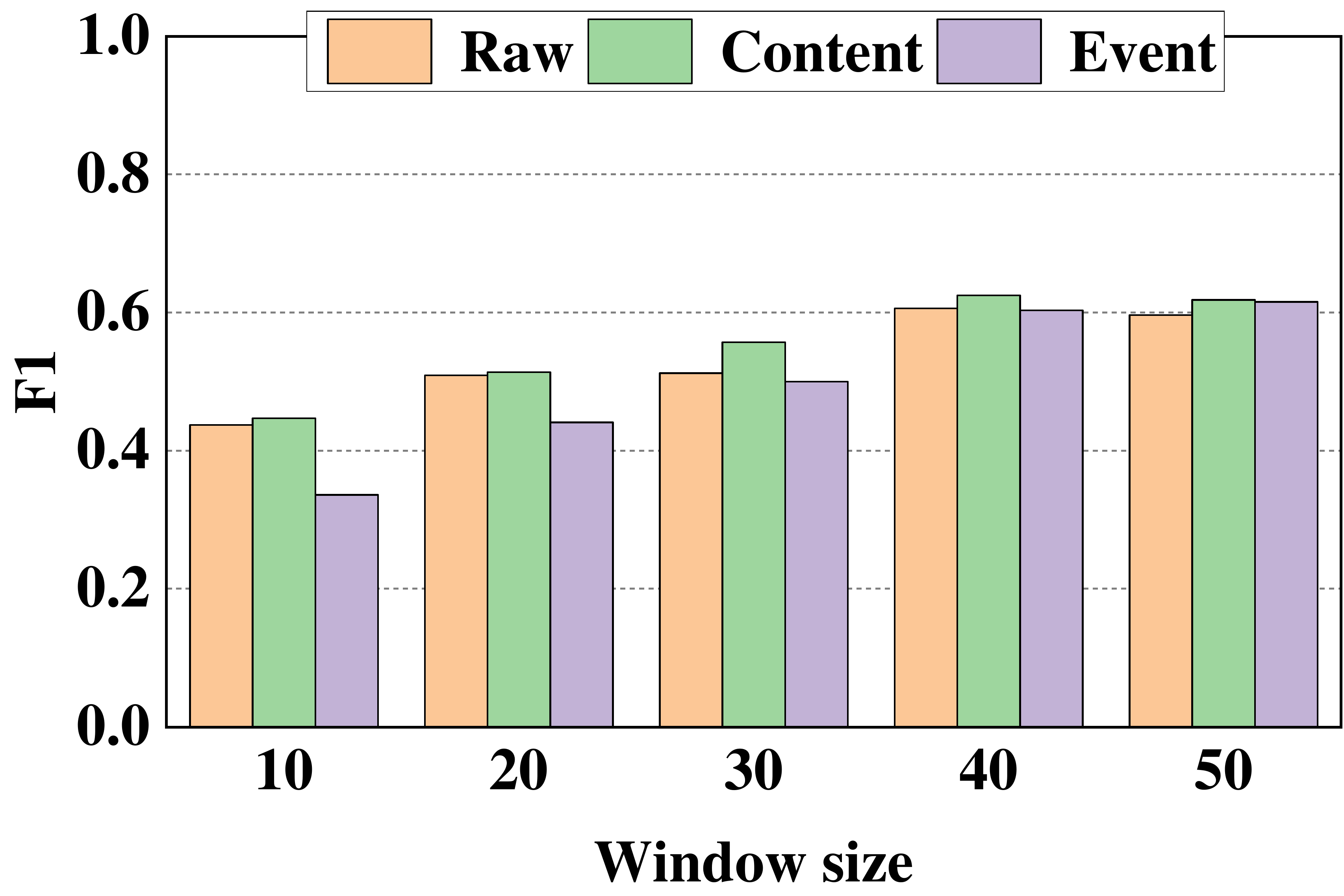}}
    \subfigure{\includegraphics[scale=0.055]{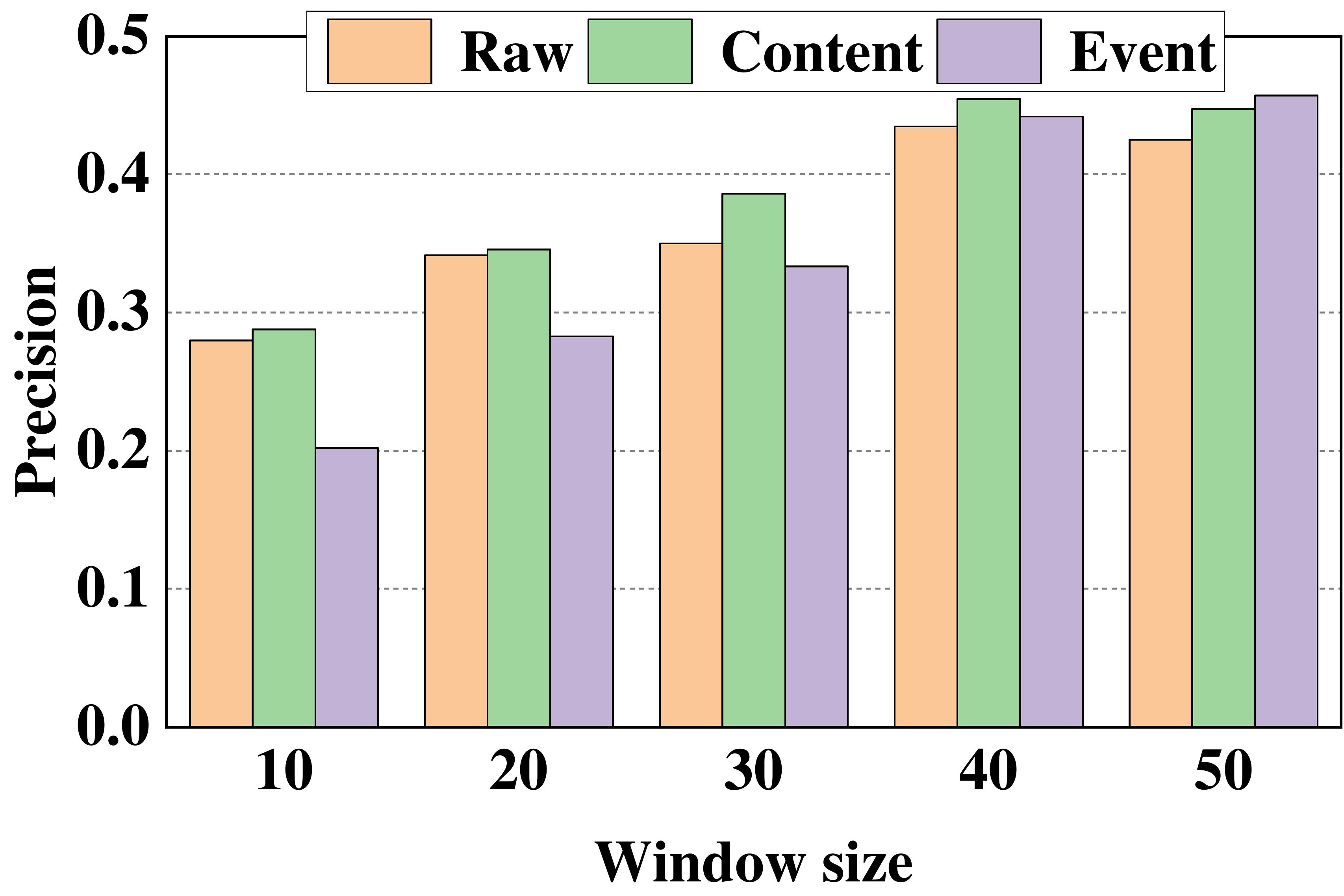}}
    \subfigure{\includegraphics[scale=0.055]{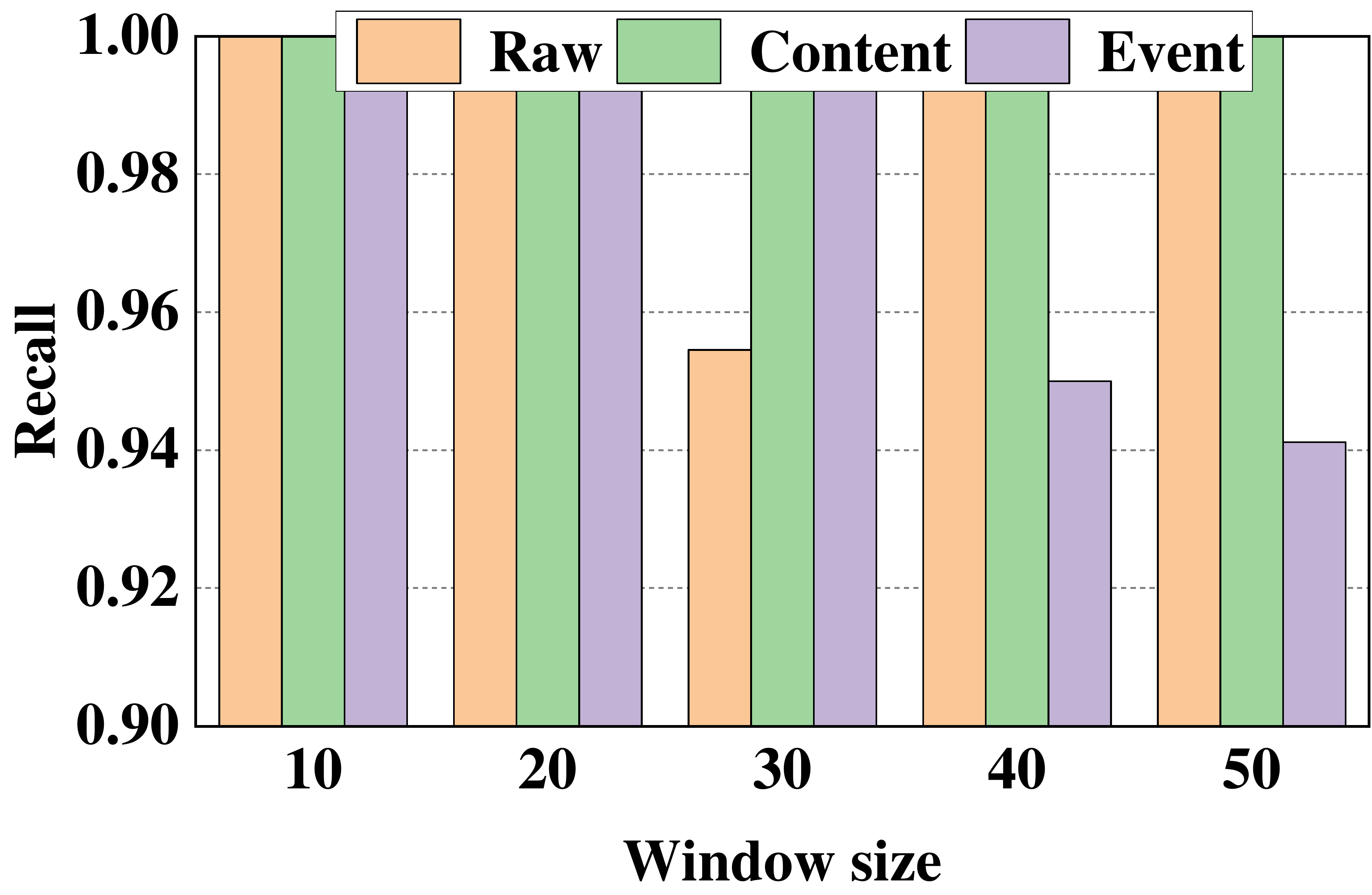}}
    \subfigure{\includegraphics[scale=0.055]{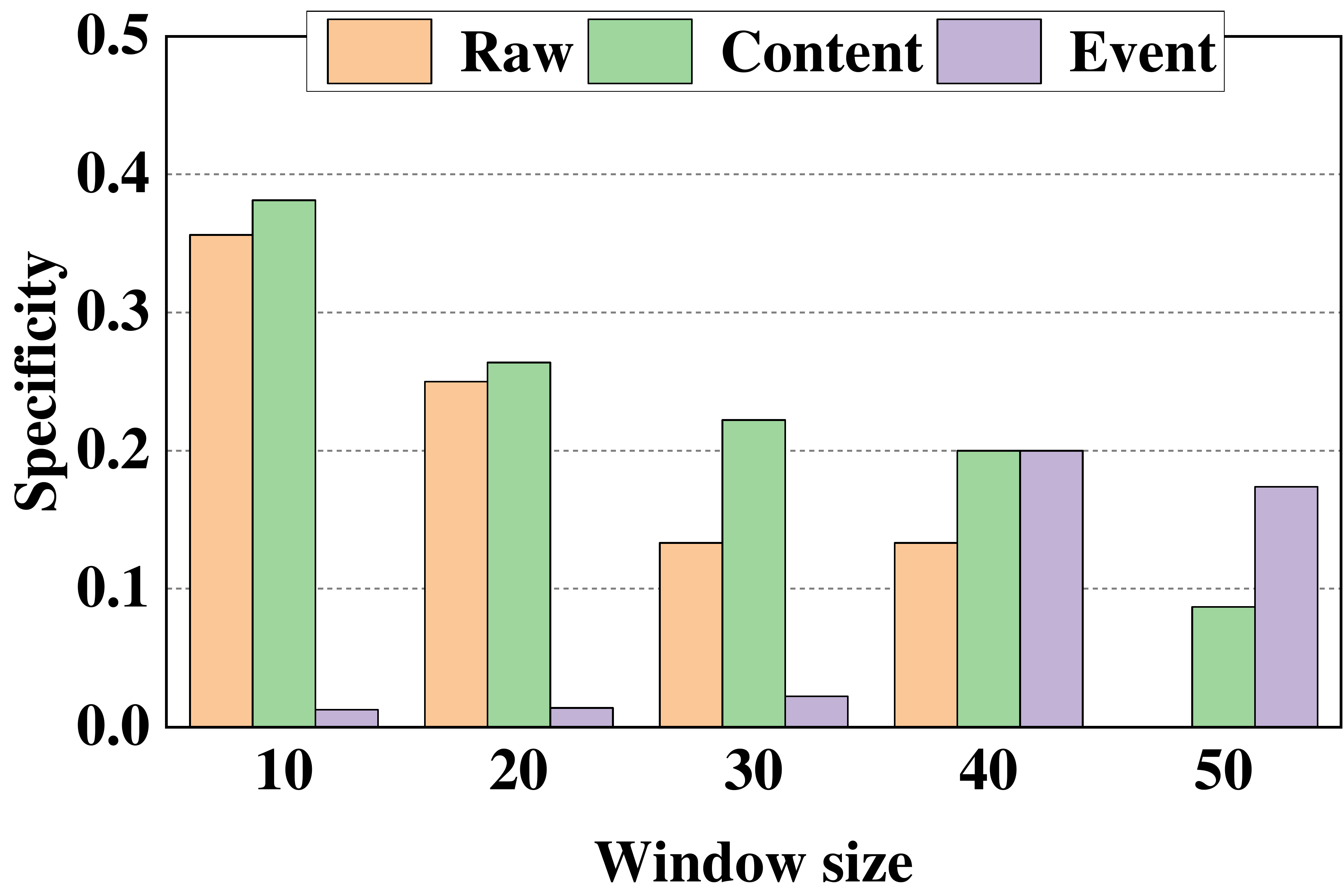}}
    \\
    \subfigure{\includegraphics[scale=0.055]{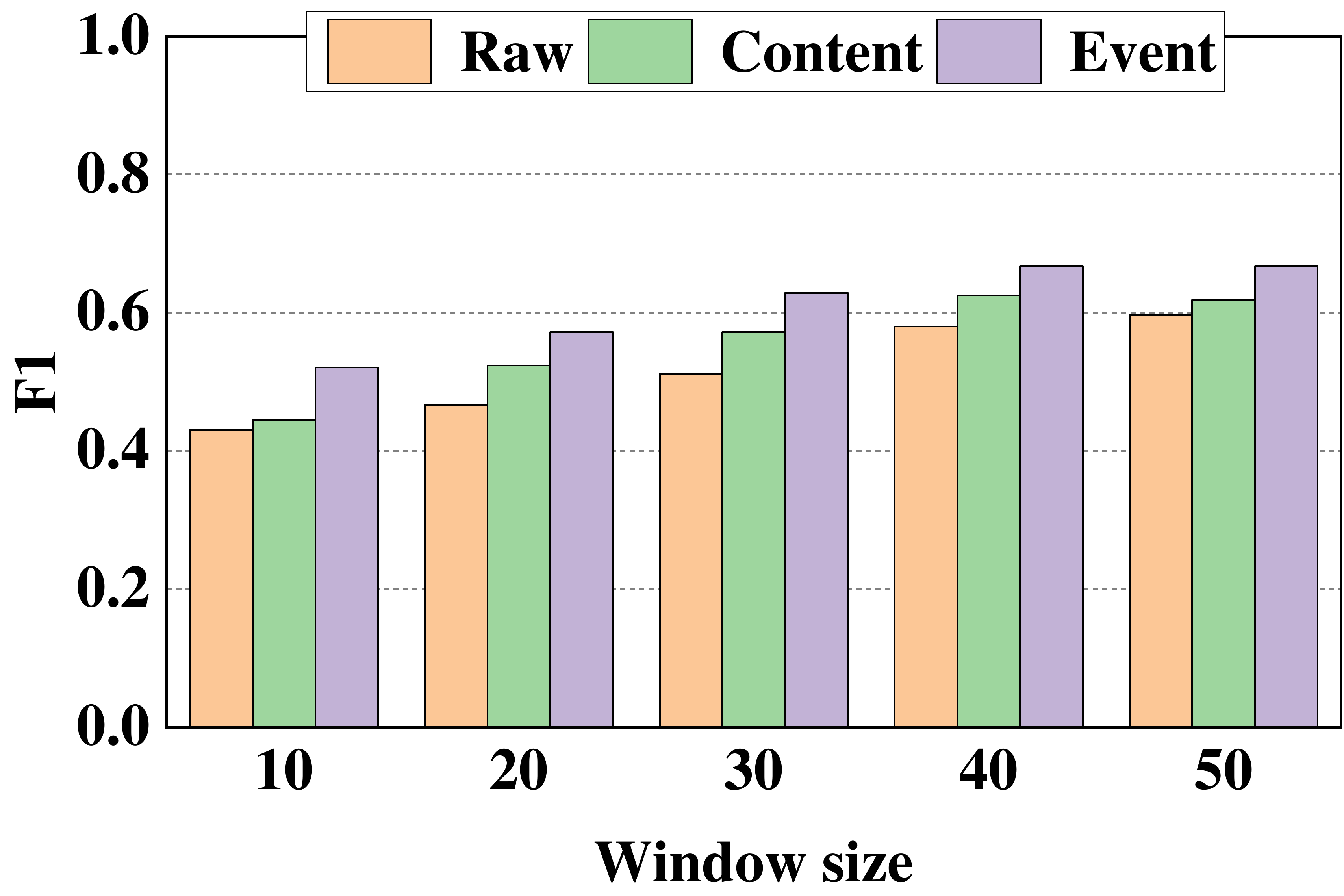}}
    \subfigure{\includegraphics[scale=0.055]{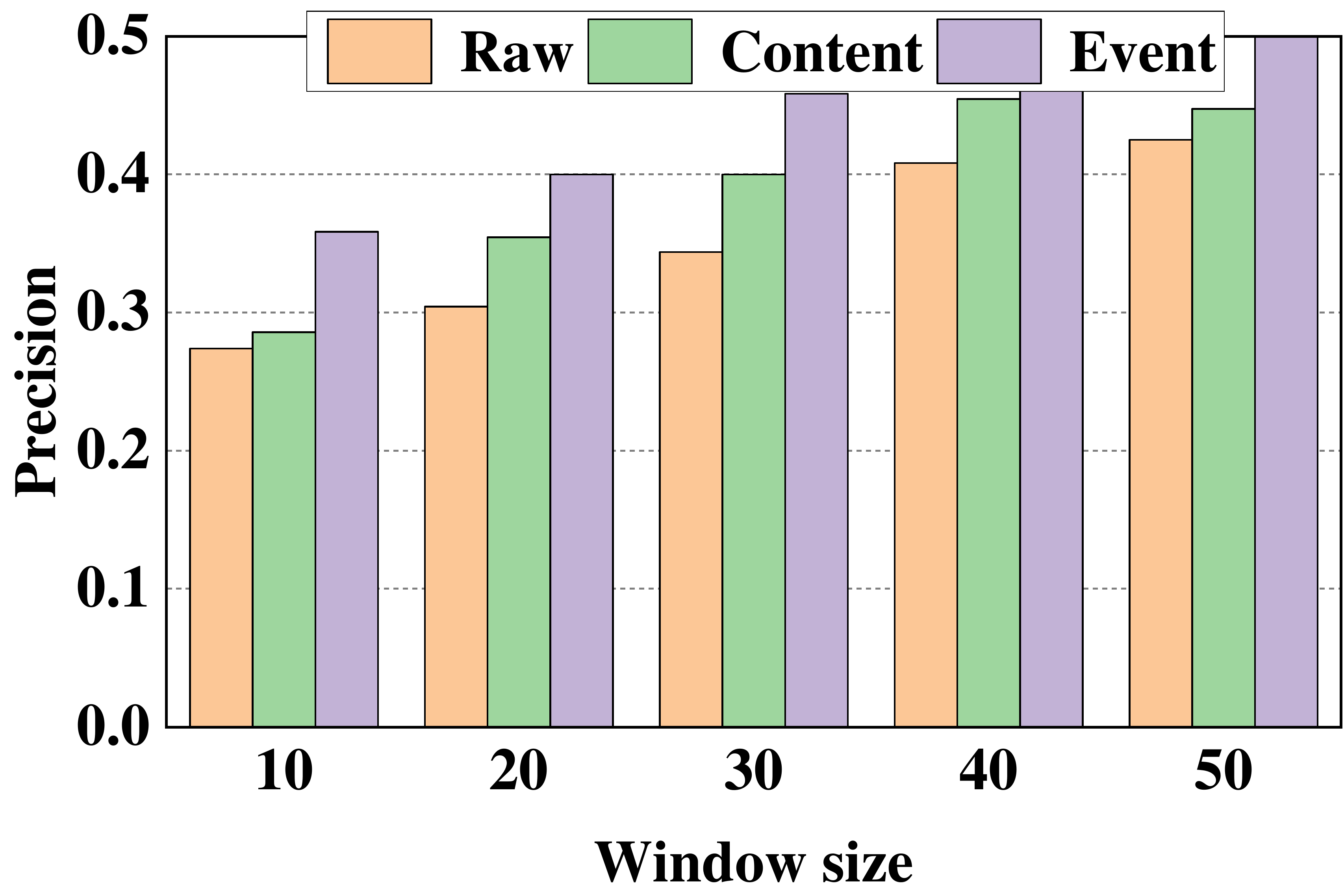}}
    \subfigure{\includegraphics[scale=0.055]{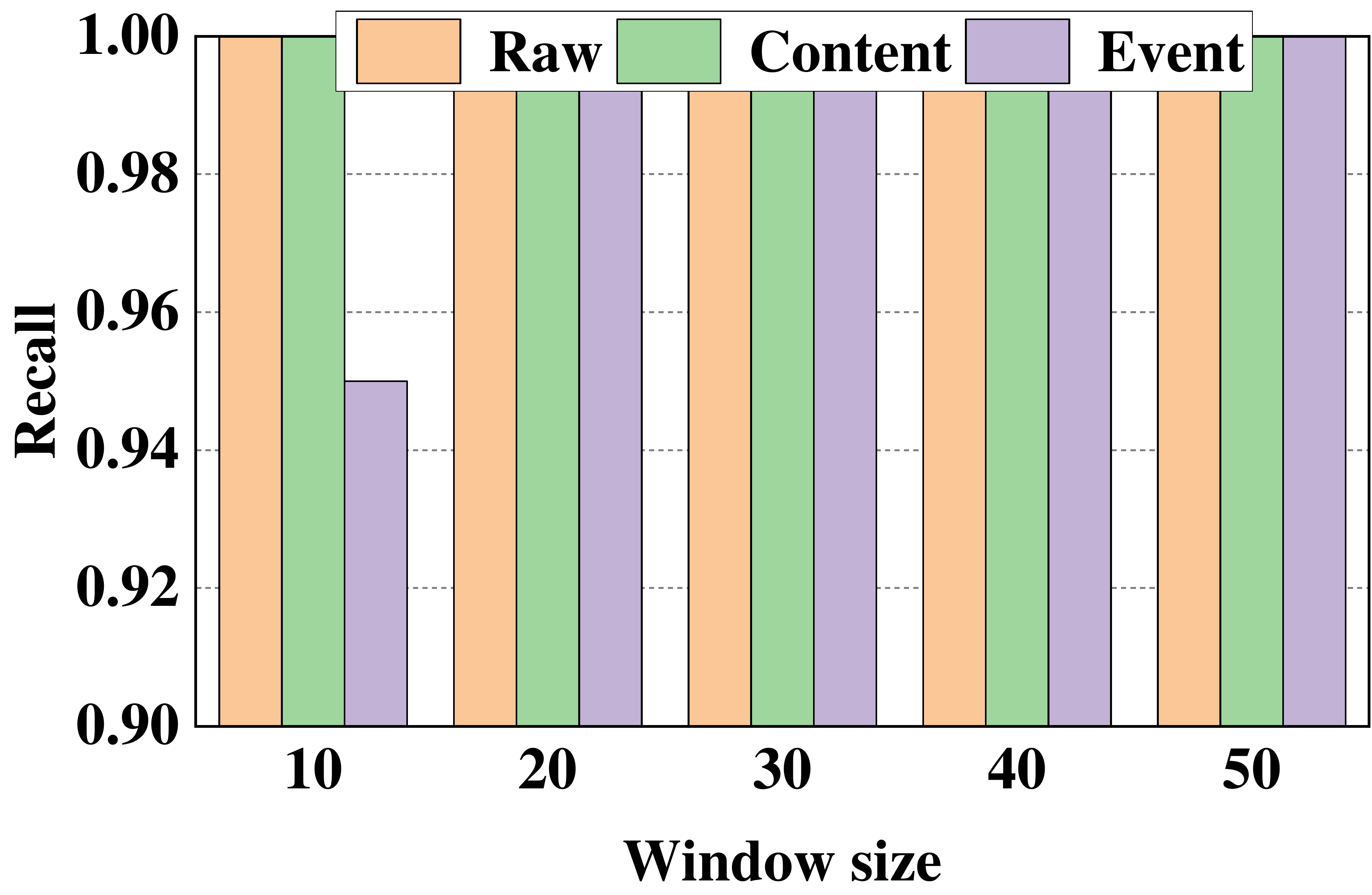}}
    \subfigure{\includegraphics[scale=0.055]{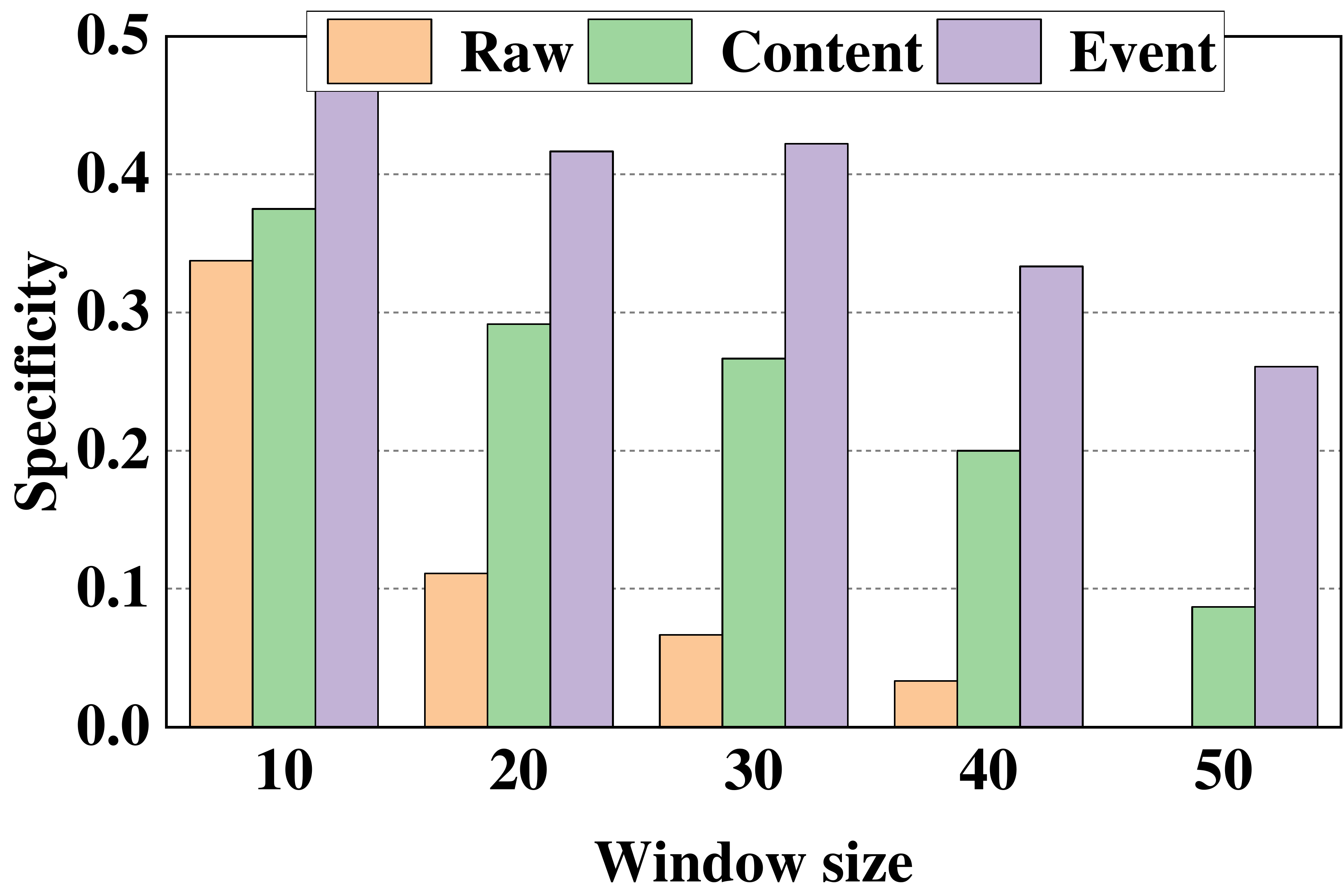}}
    
    \caption{BGL. First row: Prompt-1 (zero-shot); Second row: Prompt-2 (zero-shot); Third row: Prompt-1 (few-shot); Last row: Prompt-2 (few-shot).}
    \label{fig:bgl-seq}
\end{figure*}

\begin{figure*}[!htb]
    \centering
    \subfigure{\includegraphics[scale=0.055]{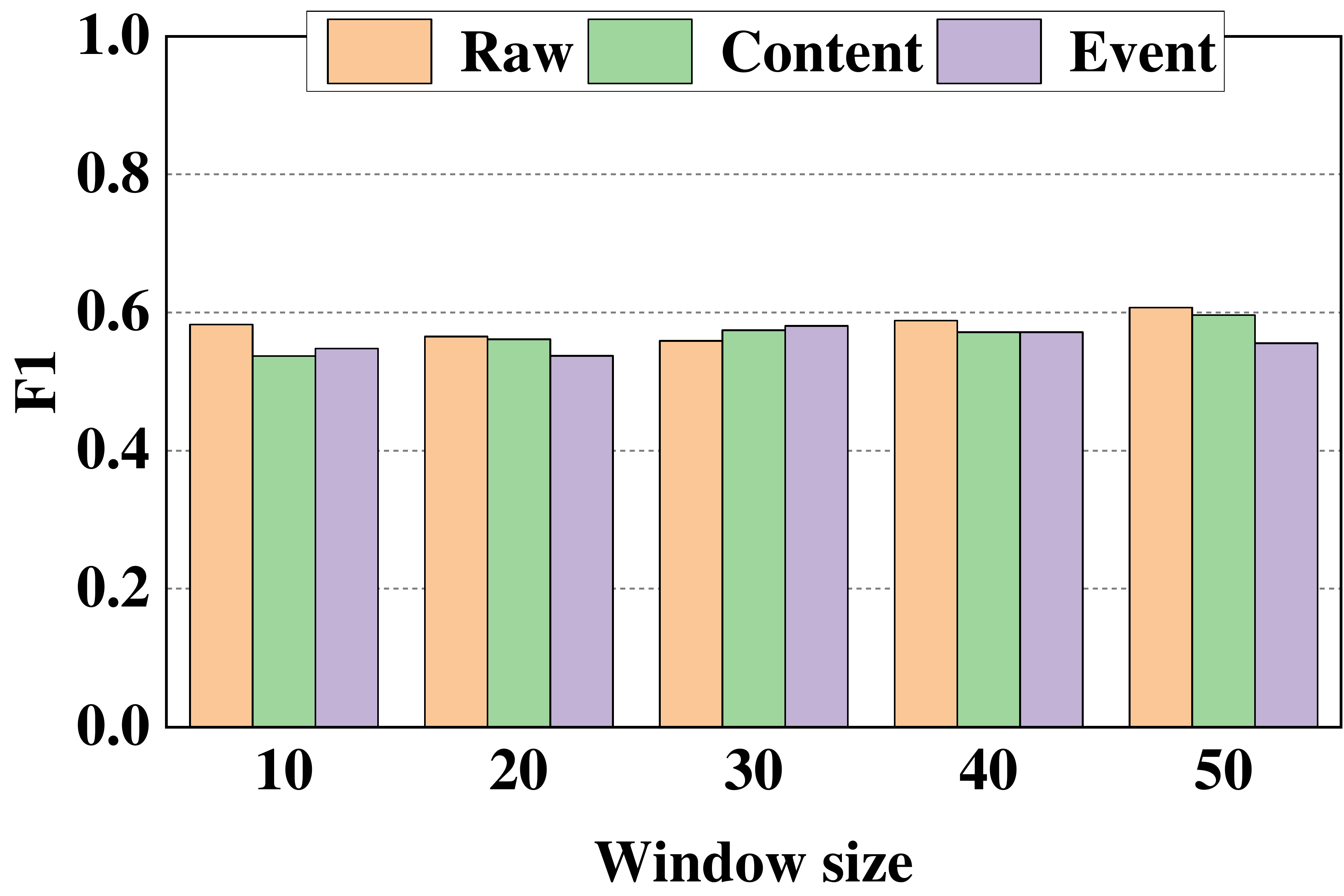}}
    \subfigure{\includegraphics[scale=0.055]{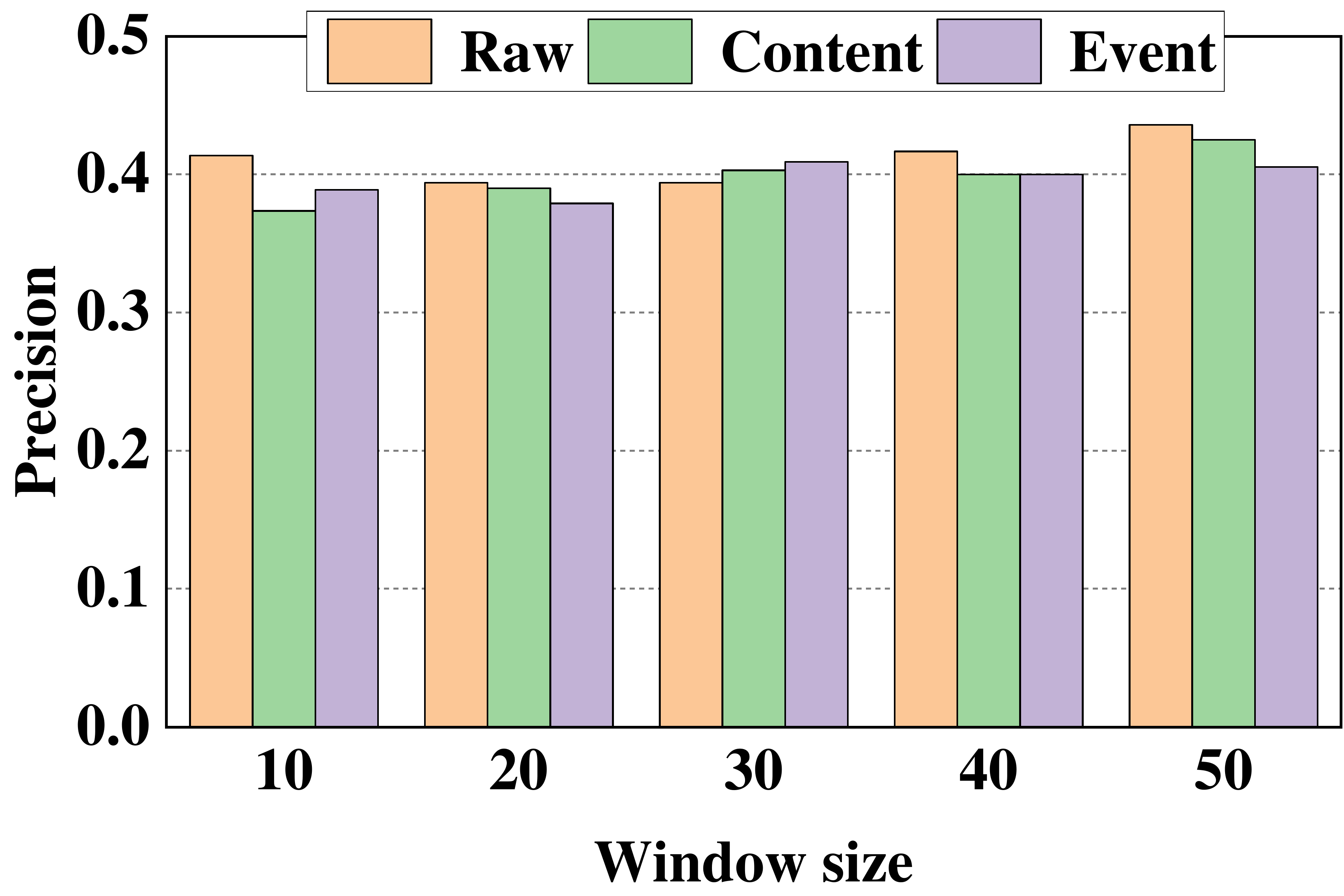}}
    \subfigure{\includegraphics[scale=0.055]{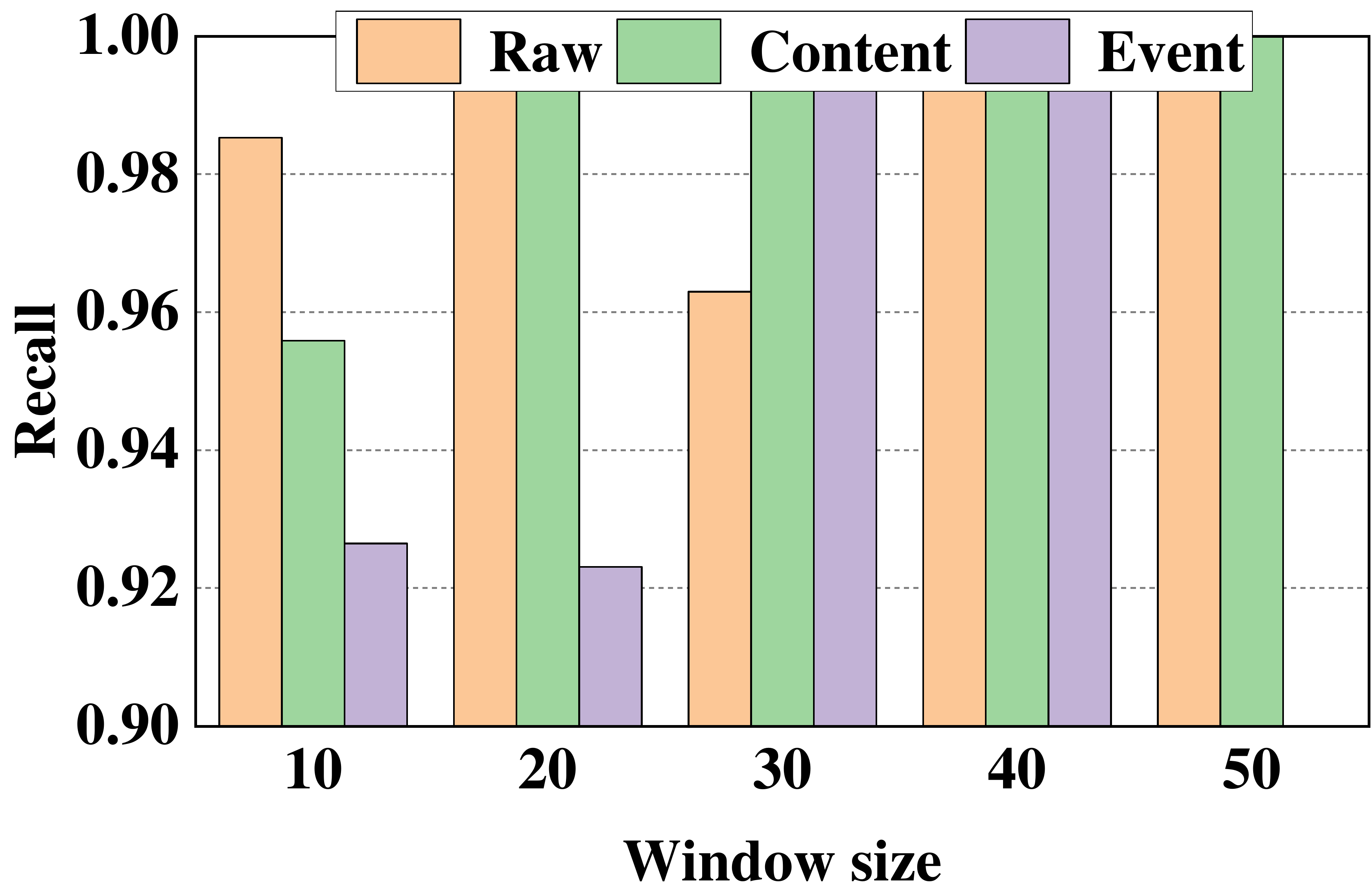}}
    \subfigure*{\includegraphics[scale=0.055]{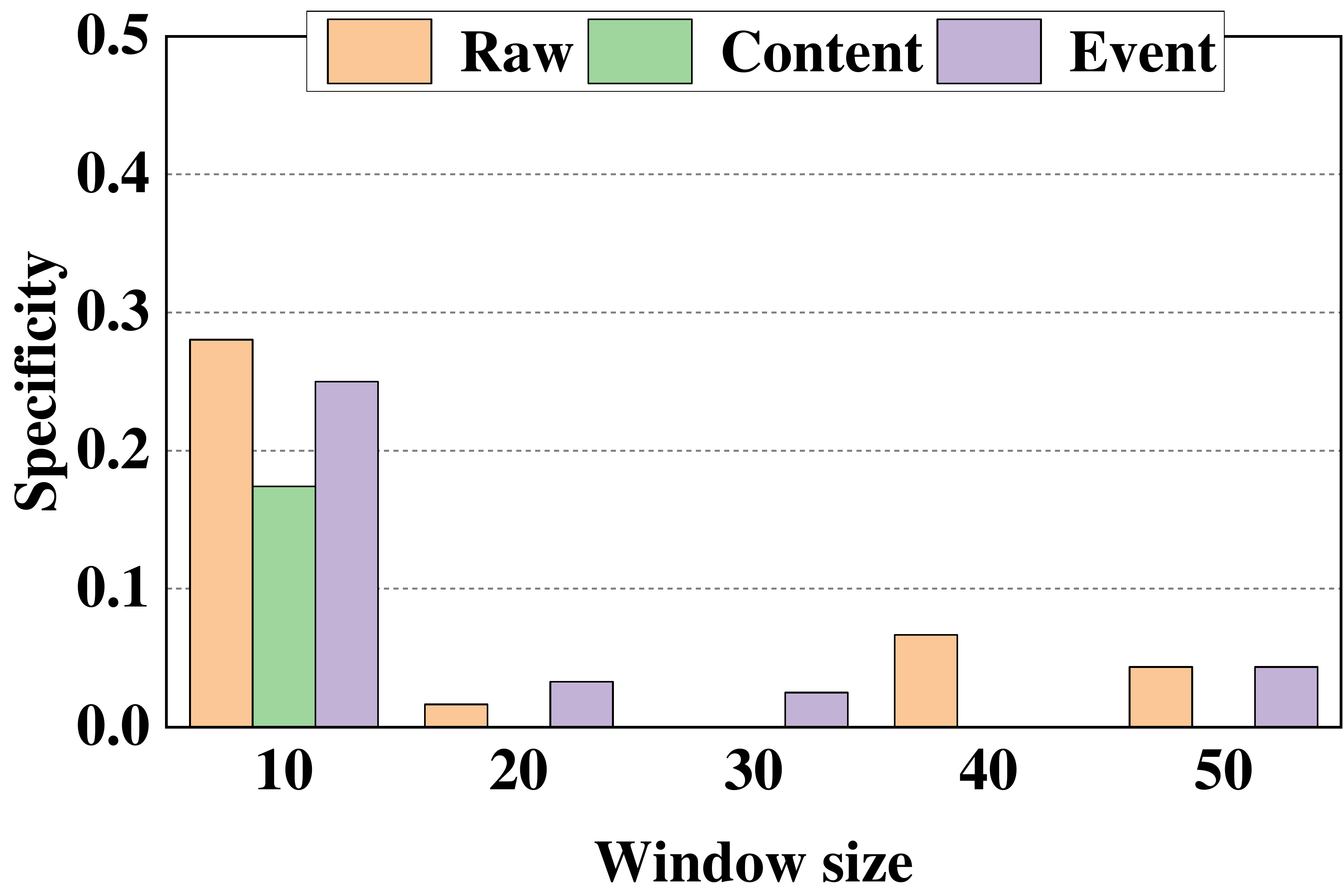}}
    \\
    \subfigure{\includegraphics[scale=0.055]{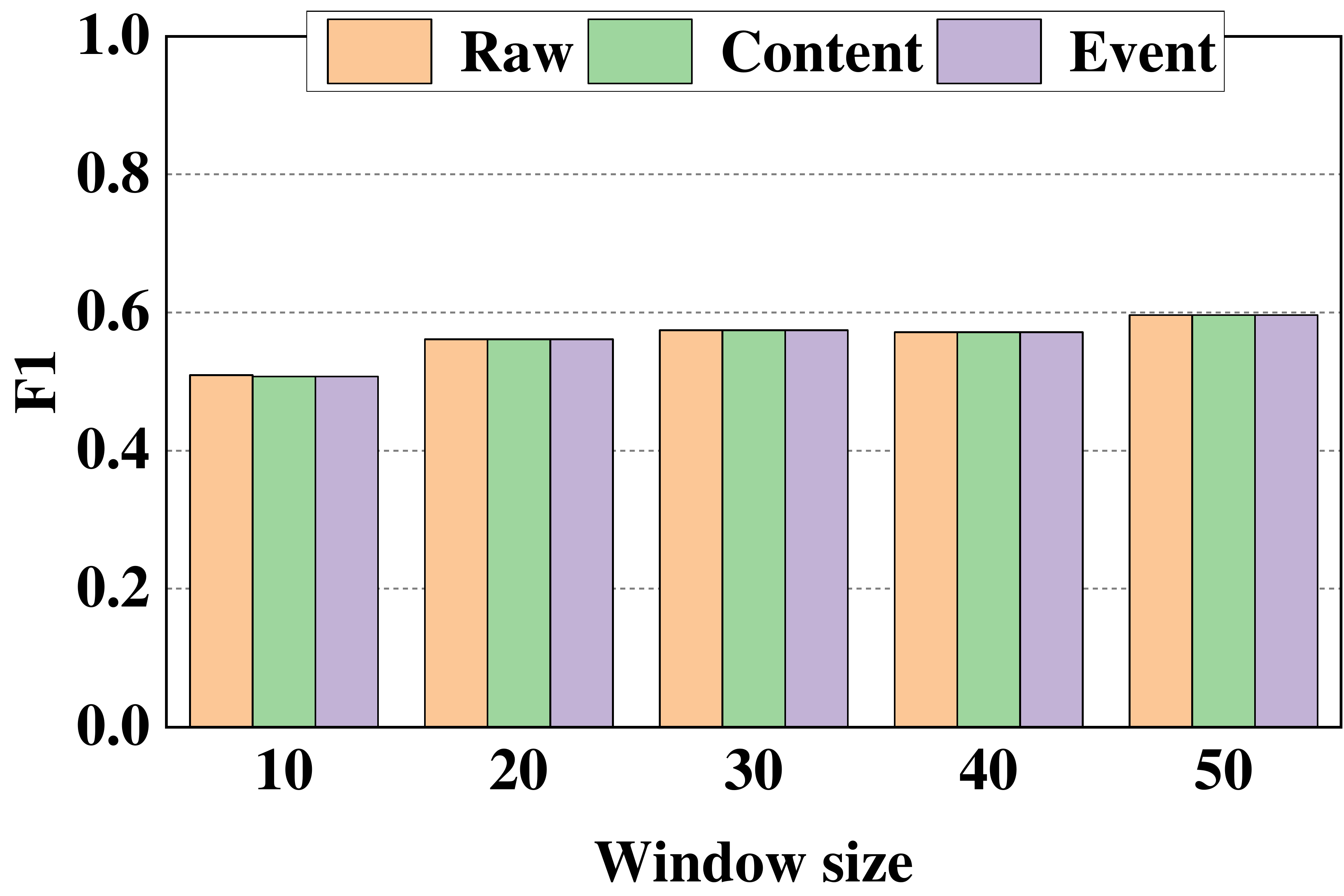}}
    \subfigure{\includegraphics[scale=0.055]{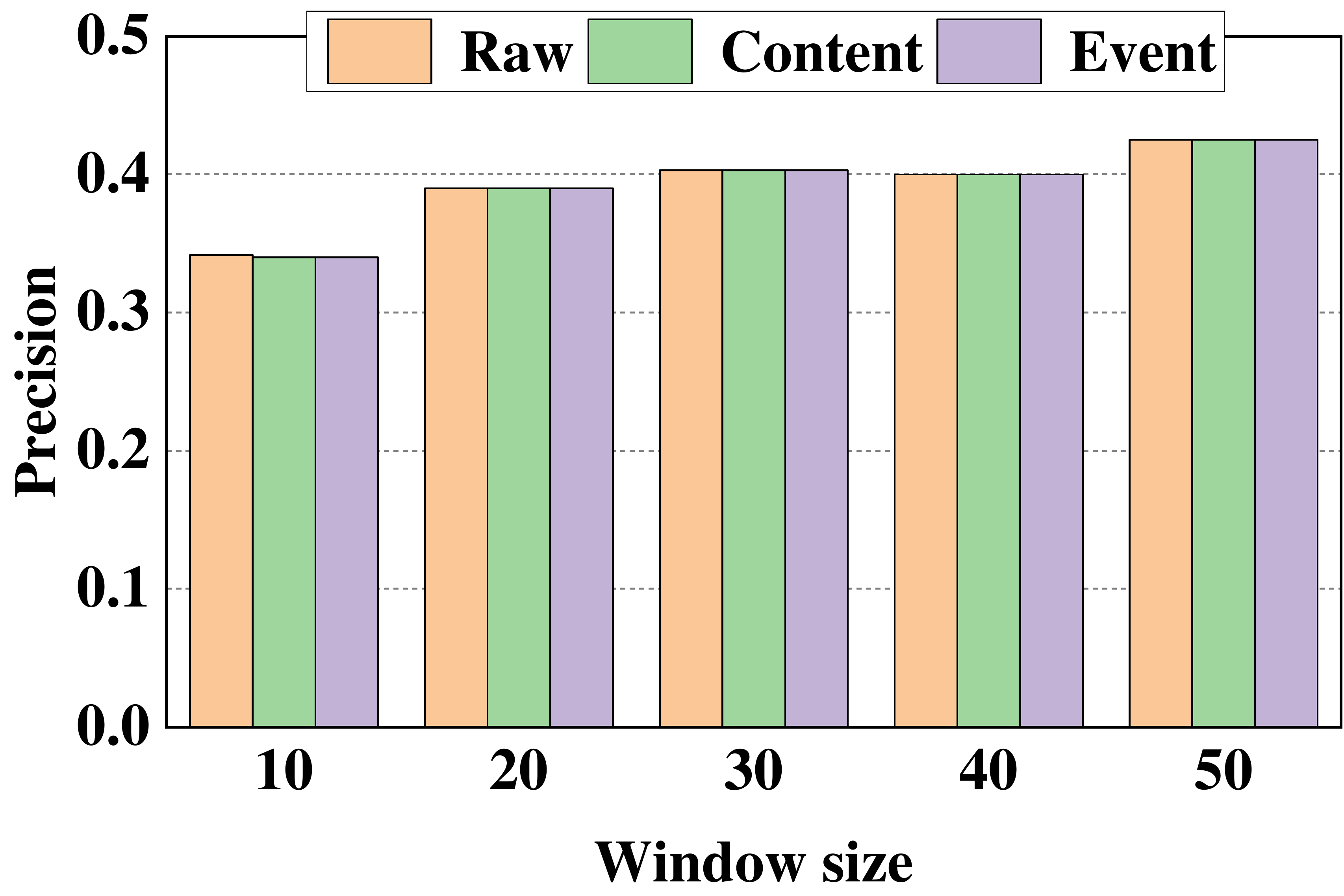}}
    \subfigure{\includegraphics[scale=0.055]{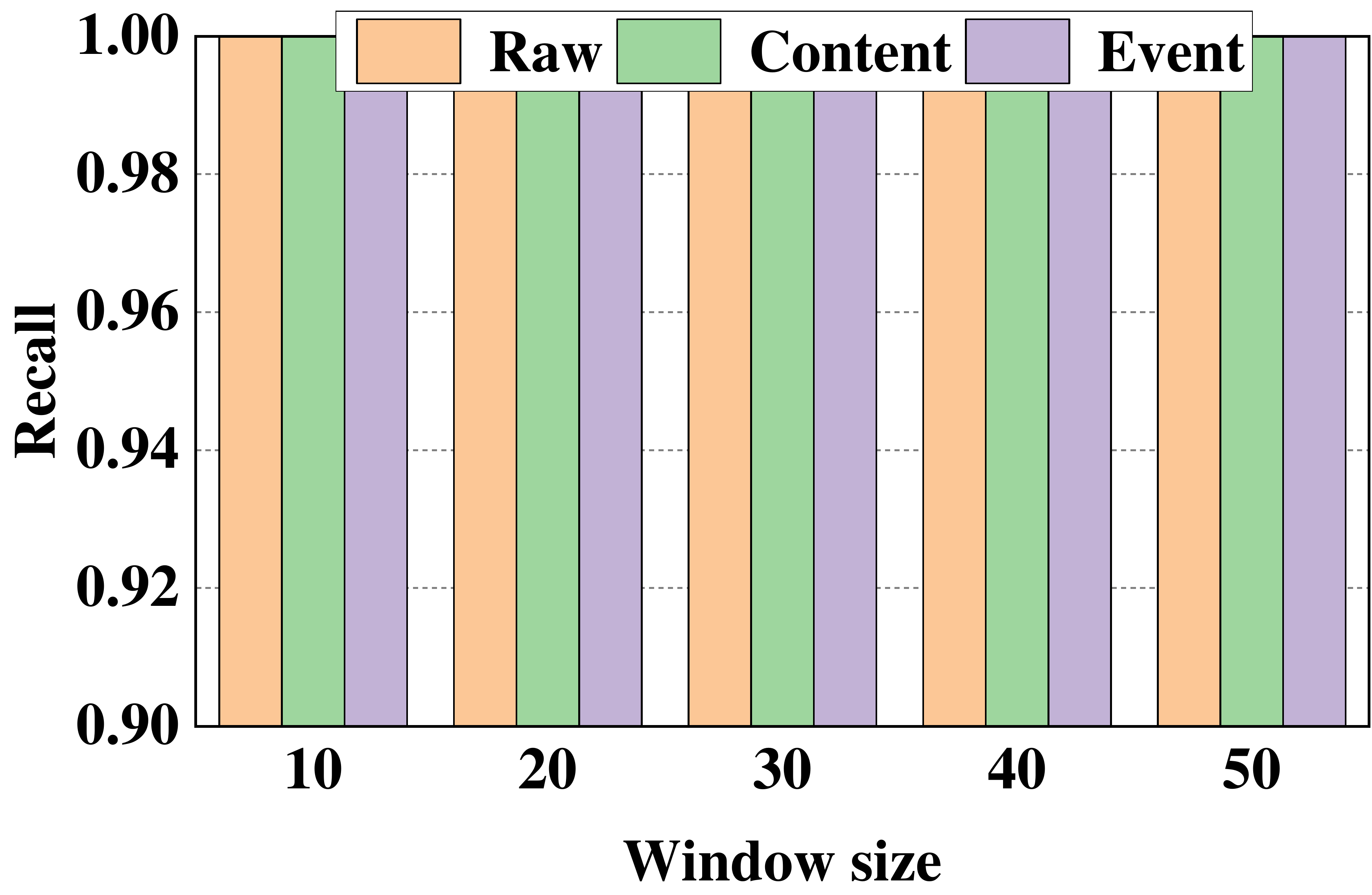}}
    \subfigure{\includegraphics[scale=0.055]{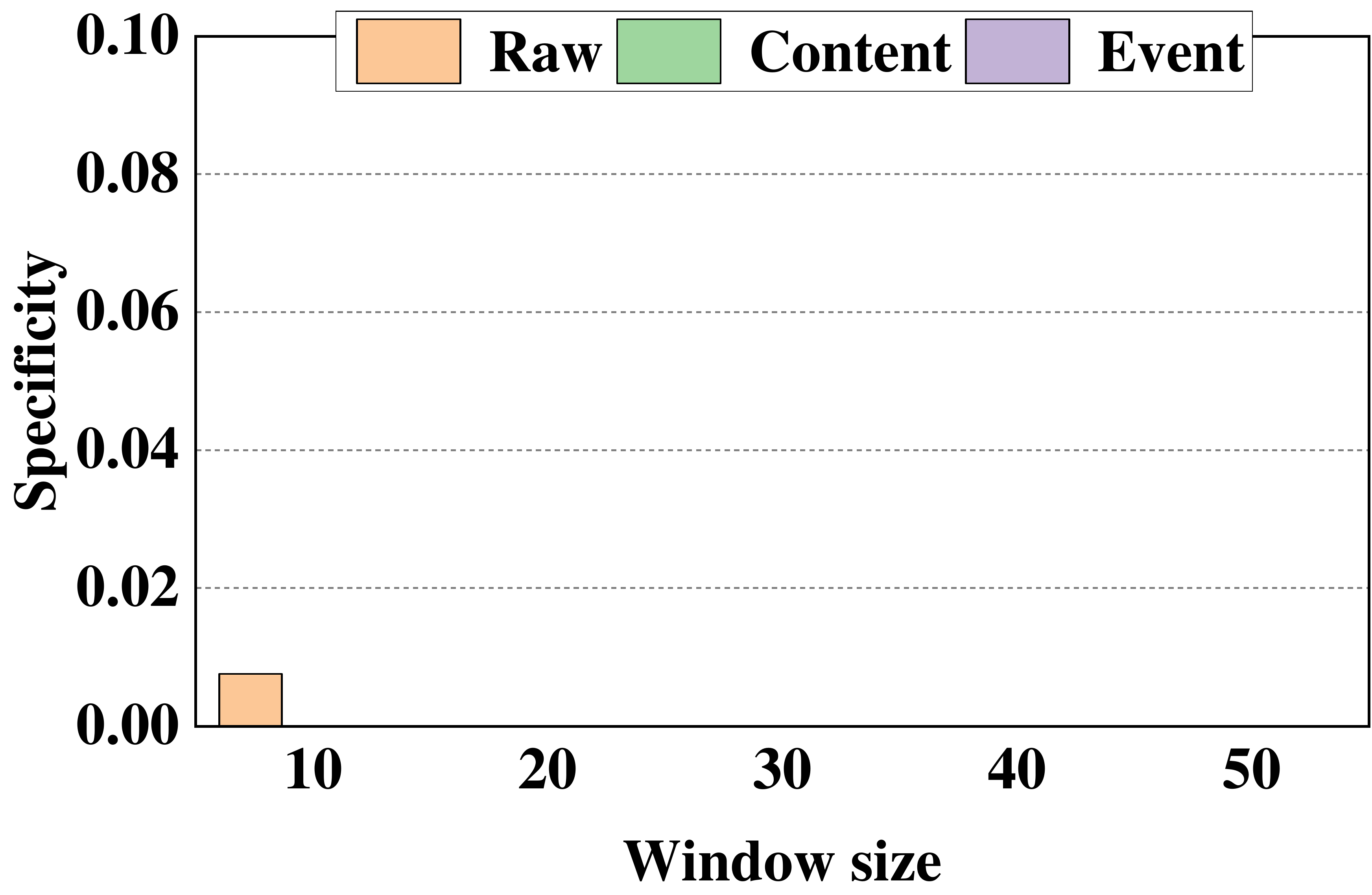}}
    \\
    \subfigure{\includegraphics[scale=0.055]{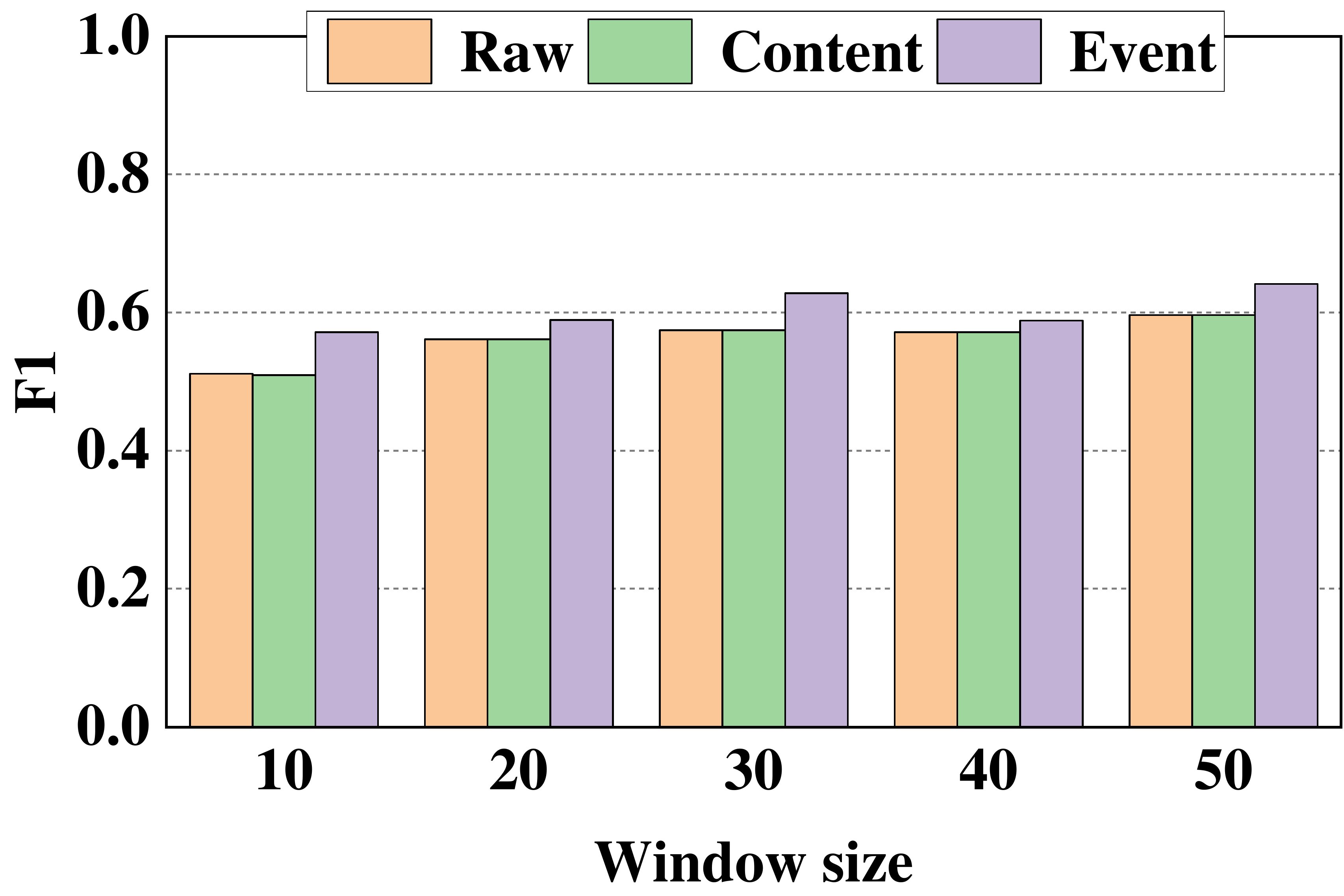}}
    \subfigure{\includegraphics[scale=0.055]{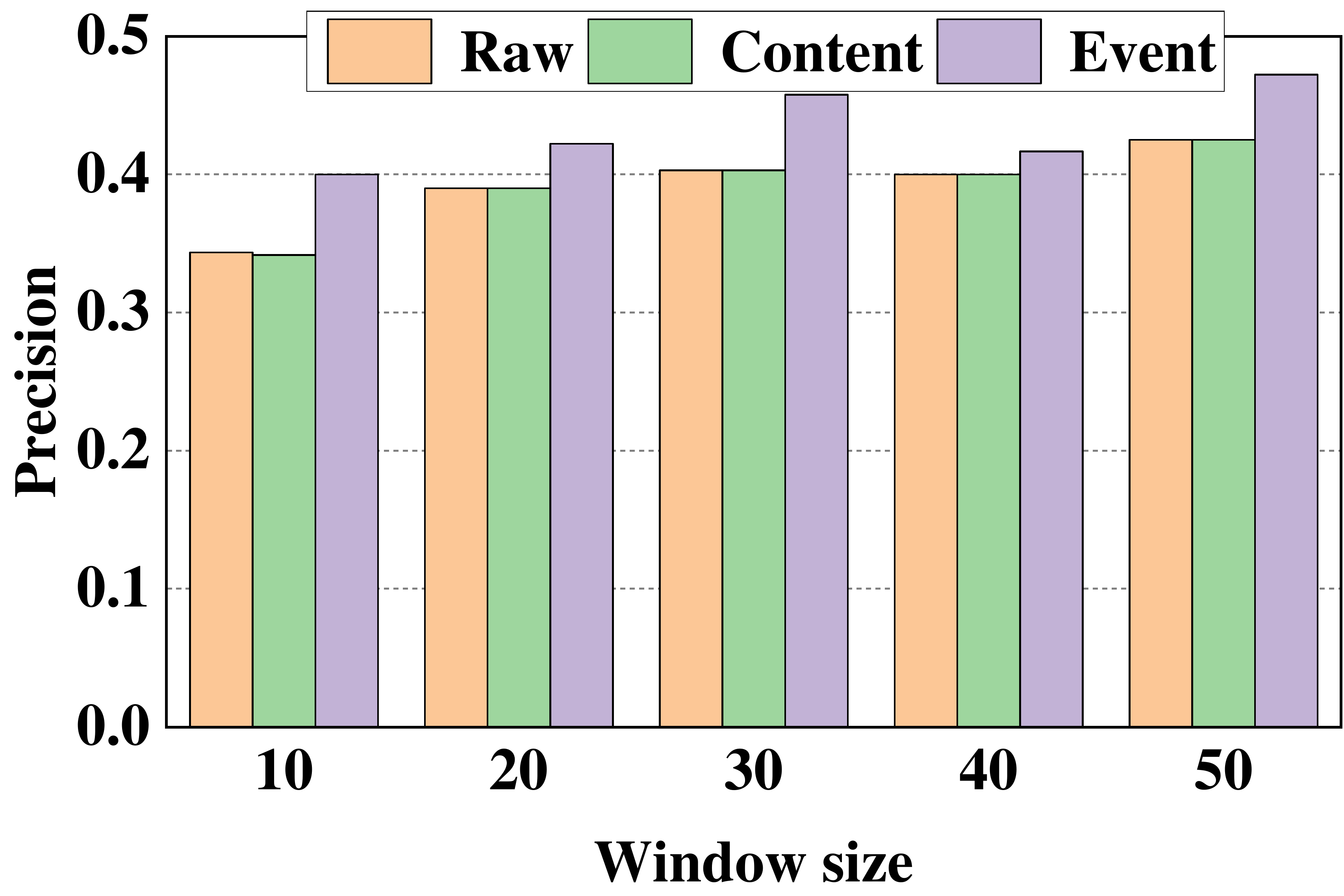}}
    \subfigure{\includegraphics[scale=0.055]{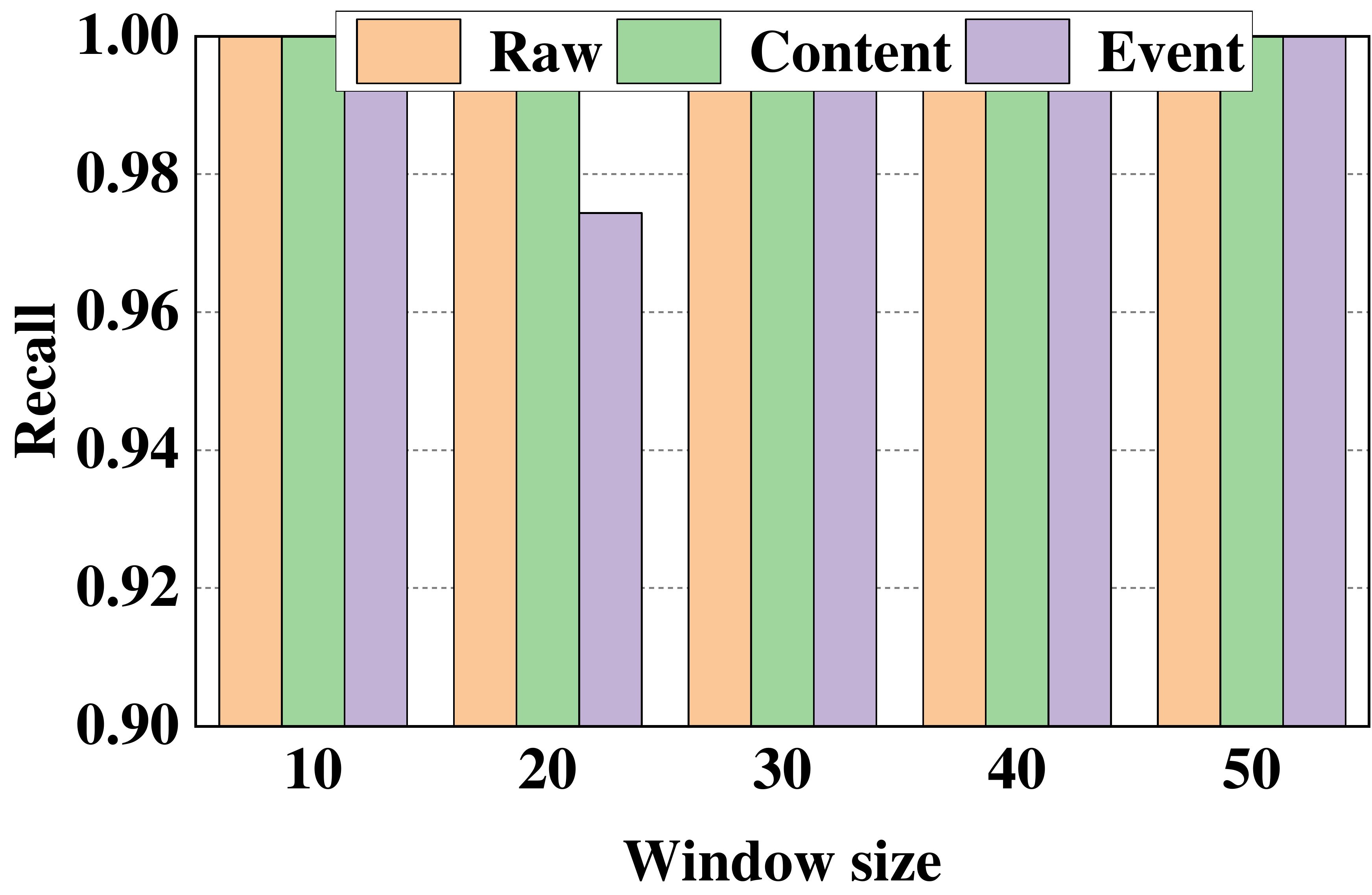}}
    \subfigure{\includegraphics[scale=0.055]{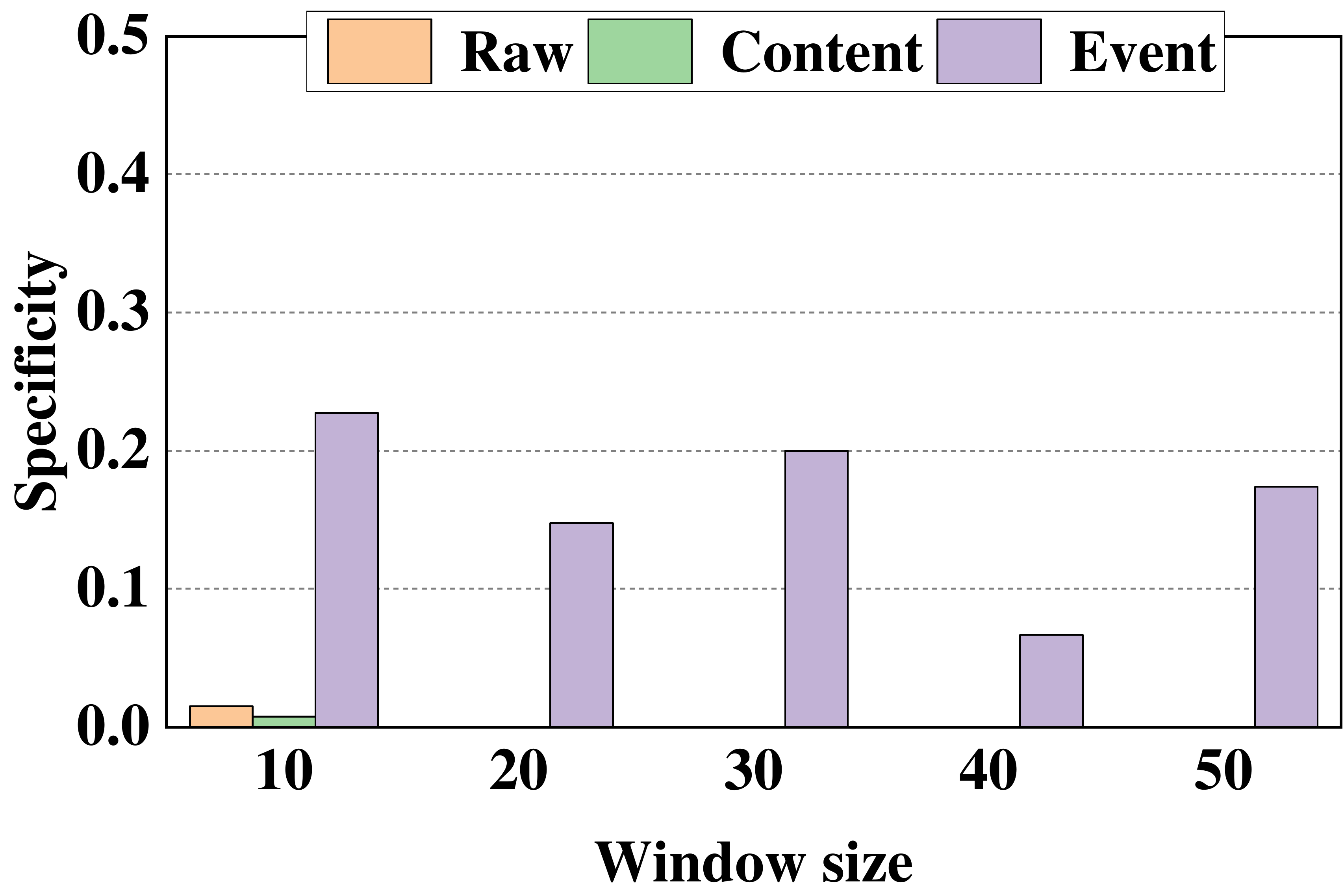}}
    \\
    \subfigure{\includegraphics[scale=0.055]{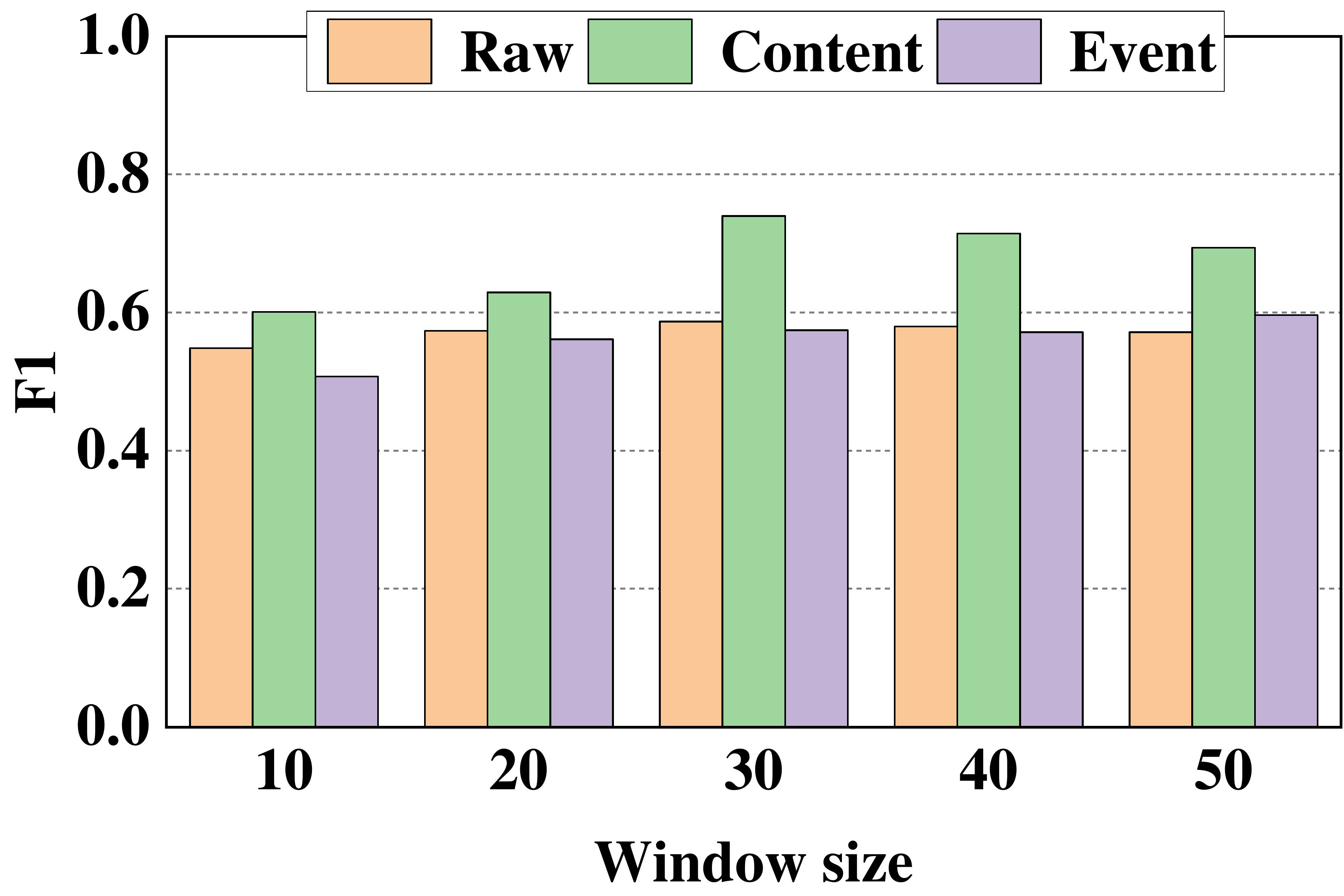}}
    \subfigure{\includegraphics[scale=0.055]{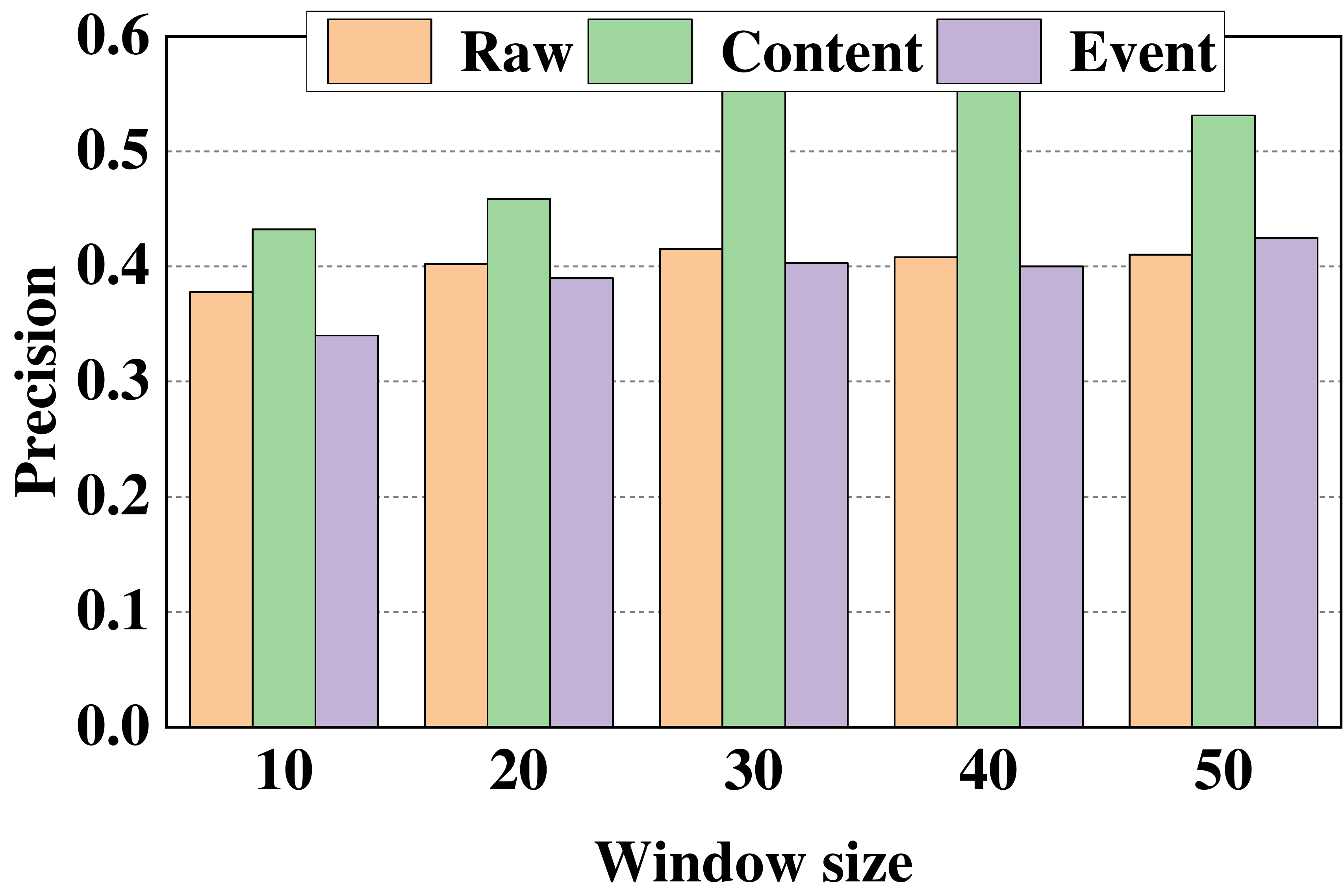}}
    \subfigure{\includegraphics[scale=0.055]{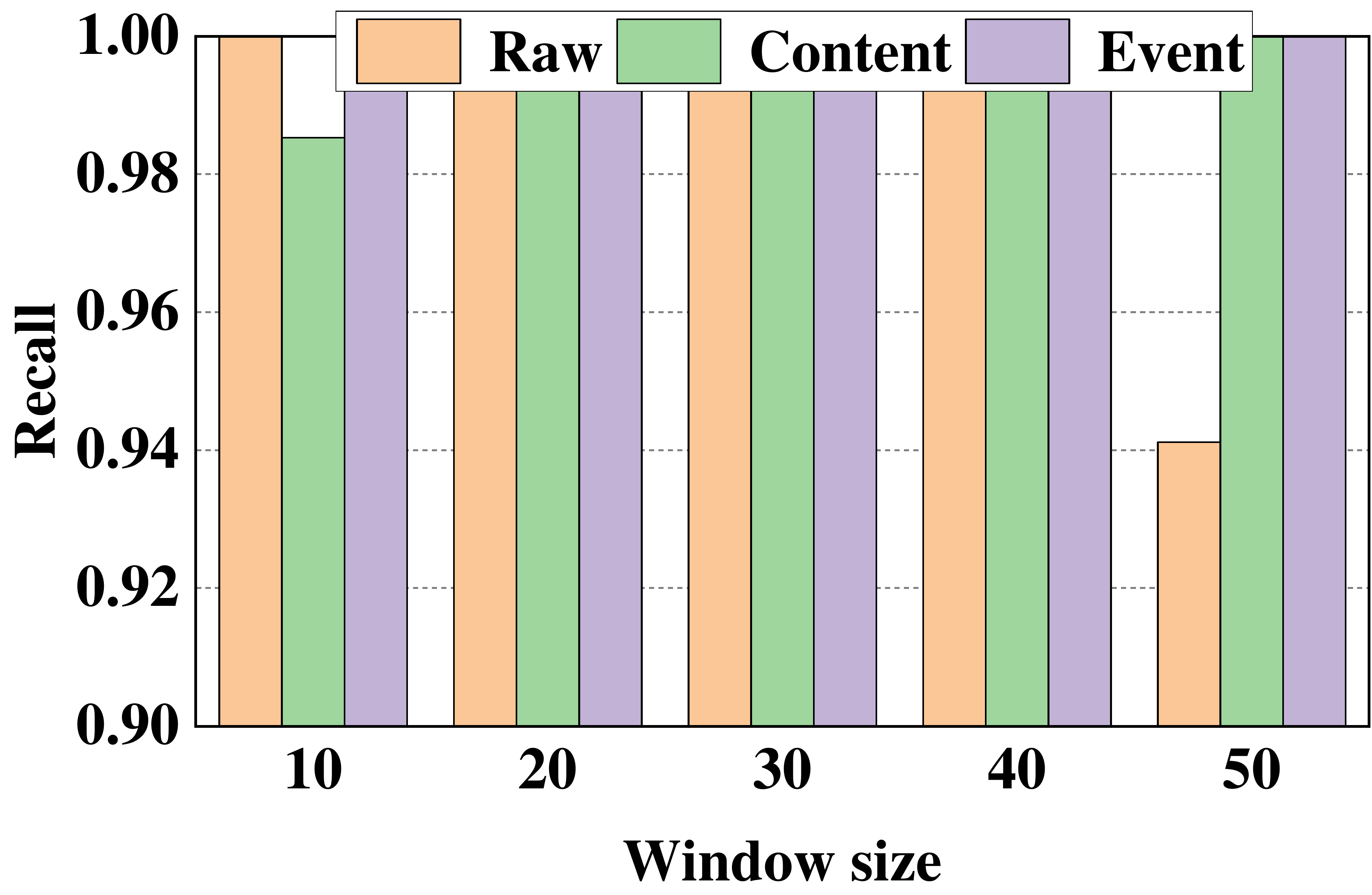}}
    \subfigure{\includegraphics[scale=0.055]{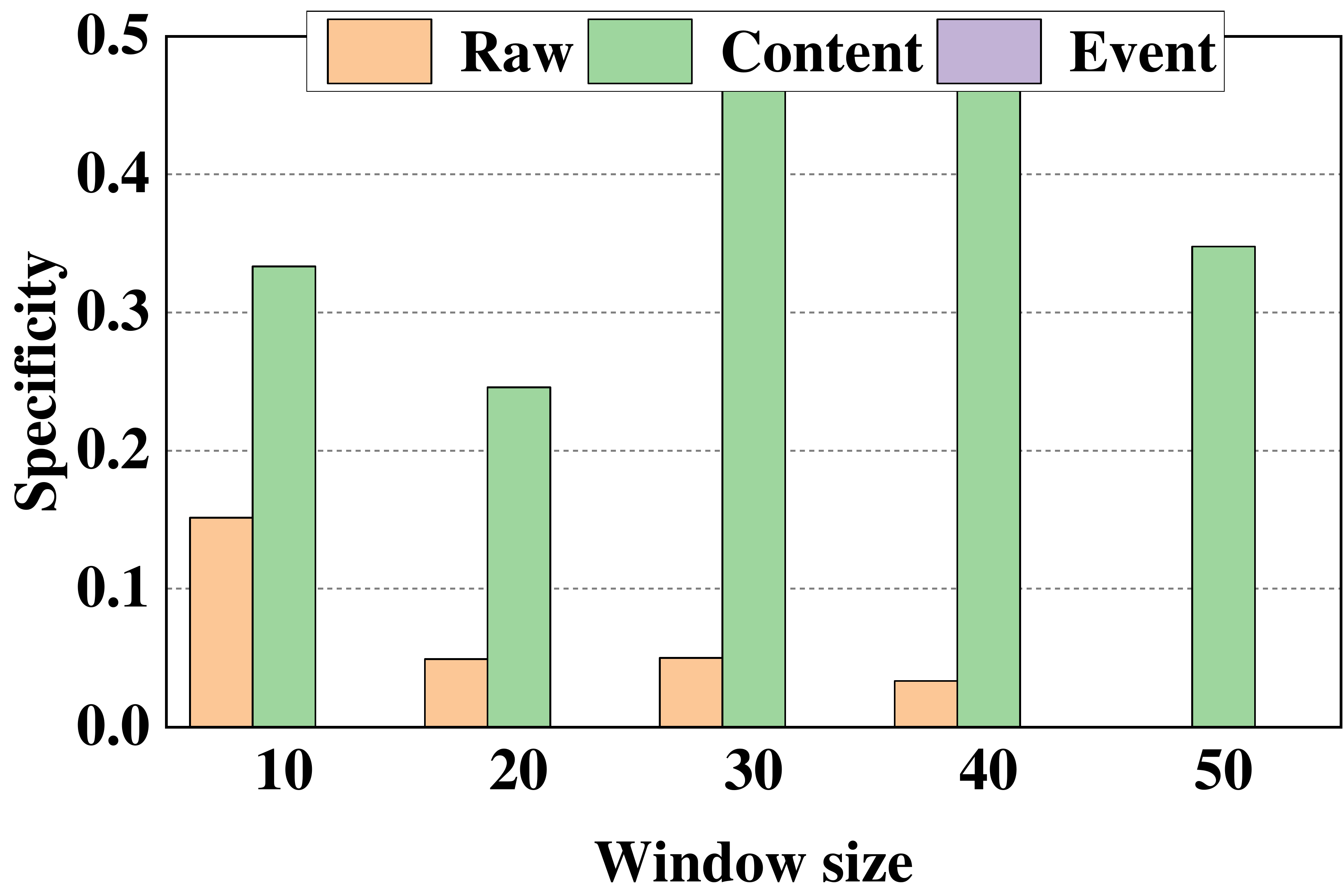}}
    
    \caption{Spirit. First row: Prompt-1 (zero-shot); Second row: Prompt-2 (zero-shot); Third row: Prompt-1 (few-shot); Last row: Prompt-2 (few-shot).}
    \label{fig:spirit-seq}
\end{figure*}

More experimental results on both BGL and Spirit datasets are shown in Figure \ref{fig:bgl-seq} and Figure \ref{fig:spirit-seq}. The window size is increased from 10 to 50 and all three types of input sequences are evaluated. The \textit{gpt-3.5-turbo} model is used in our experiments. Our source code and detailed experimental data will be made available with the publication of this work.

\end{document}